\documentclass[10pt, a4paper, twocolumn]{article}

\usepackage[utf8]{inputenc}
\usepackage[T1]{fontenc}
\usepackage[english]{babel}

\usepackage{tgheros} 

\usepackage[varqu]{zi4}
\usepackage{microtype}

\usepackage{titlesec}
\titlespacing*{\section}{0pt}{1.5ex plus 1ex minus .2ex}{0.1em}
\titlespacing*{\subsection}{0pt}{1.5ex plus 1ex minus .2ex}{0.1em}
\titlespacing*{\subsubsection}{0pt}{1.5ex plus 1ex minus .2ex}{0.1em}

\usepackage[table]{xcolor}
\usepackage{booktabs}
\usepackage{float}
\usepackage{graphicx}
\usepackage{subcaption}
\usepackage[labelfont={bf,small}, textfont={small}]{caption}

\definecolor{acqua_celeste}{RGB}{0, 160, 175}

\usepackage{amsmath}
\usepackage{amssymb}
\usepackage{mathtools}
\usepackage{amsthm}
\usepackage{bbm}        
\usepackage{pifont}
\usepackage{multirow}
\usepackage{arydshln}
\usepackage{array}
\usepackage{makecell}
\usepackage{wrapfig}
\usepackage{tabularx}

\usepackage[most]{tcolorbox}
\tcbuselibrary{breakable}
\newtcolorbox{promptbox}{colback=gray!10, colframe=gray!30,
  sharp corners, boxrule=0.4pt, fontupper=\footnotesize,
  breakable, enhanced jigsaw}

\definecolor{pgreen}{rgb}{0.13, 0.55, 0.13}
\definecolor{pred}{rgb}{0.8, 0.13, 0.13}
\definecolor{diversity}{HTML}{D4E1F5}
\definecolor{recognizability}{HTML}{FFE6CC}

\newcolumntype{Y}{>{\centering\arraybackslash}X}

\theoremstyle{plain}

\theoremstyle{definition}

\theoremstyle{remark}

\usepackage[left=1.5cm, right=1.5cm, top=1.5cm, bottom=2cm]{geometry}
\usepackage[numbers, sort&compress]{natbib}
\usepackage{hyperref}
\usepackage[capitalize,noabbrev]{cleveref}

\hypersetup{
    colorlinks=true,
    linkcolor=black,
    urlcolor=acqua_celeste,
    citecolor=acqua_celeste
}

\usepackage{fancyhdr}
\usepackage{authblk}

\title{\huge\bfseries\vspace{-1em} A Structured Benchmark for Text-Guided Anomaly Detection: When Language Stops Conditioning the Decision}

\author[1]{Stefano Samele\textsuperscript{*}}
\author[1]{Eugenio Lomurno\textsuperscript{*}}
\author[1]{Teodora Jovanovic}
\author[1]{Sanjay Shivakumar Manohar}
\author[2]{Alberto Crivellaro}
\author[1]{Matteo Matteucci}

\affil[1]{Politecnico di Milano, AIRLab, Milan, Italy}
\affil[2]{S\&H -- Software \& Hardware, Milan, Italy}

\date{}

\begin{document}

\twocolumn[
  \begin{@twocolumnfalse}
    \maketitle
    \vspace{-2em}

    \begin{abstract}
        \setlength{\parindent}{0pt}
        \setlength{\parskip}{4pt}
        \itshape
        \noindent Industrial anomaly detection has historically been a unimodal task. Recent multimodal vision-language models have produced a new class of systems that admit textual input alongside the image and are presented as enabling text-guided zero- and few-shot inspection. Yet these methods are evaluated with protocols inherited from unimodal visual benchmarks that hold the textual condition effectively constant and therefore cannot measure whether language conditions the decision; whether reported gains reflect text guidance or strong pretrained visual features remains open. We introduce Text-Guided Anomaly Detection (TGAD), a structured benchmark that progressively increases the functional role of language across three scenarios: a controlled prompt-sensitivity setting on MVTec~AD; a component-tagged extension of MVTec~AD that requires the model to restrict its assessment to an instructed part; and the new Assembled Panel Dataset (APD), a realistic industrial setting that requires both defect-type and component-location knowledge. We evaluate one representative model per paradigm: generative large vision-language, training-free discriminative, and embedding-adaptive discriminative. In all three, the textual interface conditions the decision only superficially: prompt content is absorbed unless the object noun is removed (the generative model's I-AUROC drops $97.4 \to 82.6$); component-level instructions do not constrain the global decision once defects outside the instructed part are admitted as normal ($90.3 \to 66.3$); and when the two requirements combine on APD, image-level discrimination collapses well below the MVTec level, in one case below chance ($71.2$, $50.5$, $31.5$). These results suggest that standard benchmarks overstate the text-guided capabilities of current multimodal anomaly detection systems, and that a protocol of this kind is a prerequisite for building models that can be reliably controlled through language for industrial deployment.
    \end{abstract}

    \vspace{0.5em}

    \vspace{1em}
    \hrule height 1pt
    \vspace{2em}
  \end{@twocolumnfalse}
]

{
  \renewcommand{\thefootnote}{\fnsymbol{footnote}}
  \footnotetext[1]{The authors equally contributed to this work.}
}

\section{Introduction}
\label{sec:introduction}

Anomaly detection underpins industrial visual inspection, where the system must learn a representation of normality from a finite set of defect-free examples and flag deviations at inference time \cite{chandola2009anomaly,pang2021deep}. The dominant approaches, rooted in deep learning, are reconstruction-based, embedding-based, and synthesis-based; all are unimodal, operating solely on visual input \cite{pang2021deep}. Evaluation is standardized around image-level detection and pixel-level localization on benchmarks such as MVTec~AD \cite{bergmann2021mvtec}, BTAD \cite{mishra2021vt}, WFDD \cite{chen2024unified}, and VisA \cite{zou2022spot}.

Vision-language foundation models such as CLIP \cite{radford2021learning}, and large vision-language models (LVLMs), have produced a growing class of anomaly detection methods that admit textual input alongside the image and advertise text-guided or instruction-driven behavior \cite{baltrusaitis2019multimodal, jeong2023winclip}. These methods report state-of-practice numbers on standard benchmarks and are presented as zero-shot or few-shot, plug-and-play, controllable through prompts. Whether the textual interface they expose actually conditions the decision is a separate question --- the one this paper investigates.

Existing evaluation protocols have not evolved alongside the methods they evaluate. Standard MVTec~AD evaluation reports image- and pixel-level AUROC under a fixed split that holds text content effectively constant across all test samples, and on this protocol these methods return competitive numbers. Under such a protocol, reported gains may originate from strong pretrained visual representations augmented by a generic linguistic prior, with the prompt acting as a static class label rather than a controllable instruction. This paper introduces an evaluation that addresses the question the standard protocol does not ask: whether language conditions the decision.

\begin{figure*}[t]
    \centering
    \includegraphics[width=\linewidth]{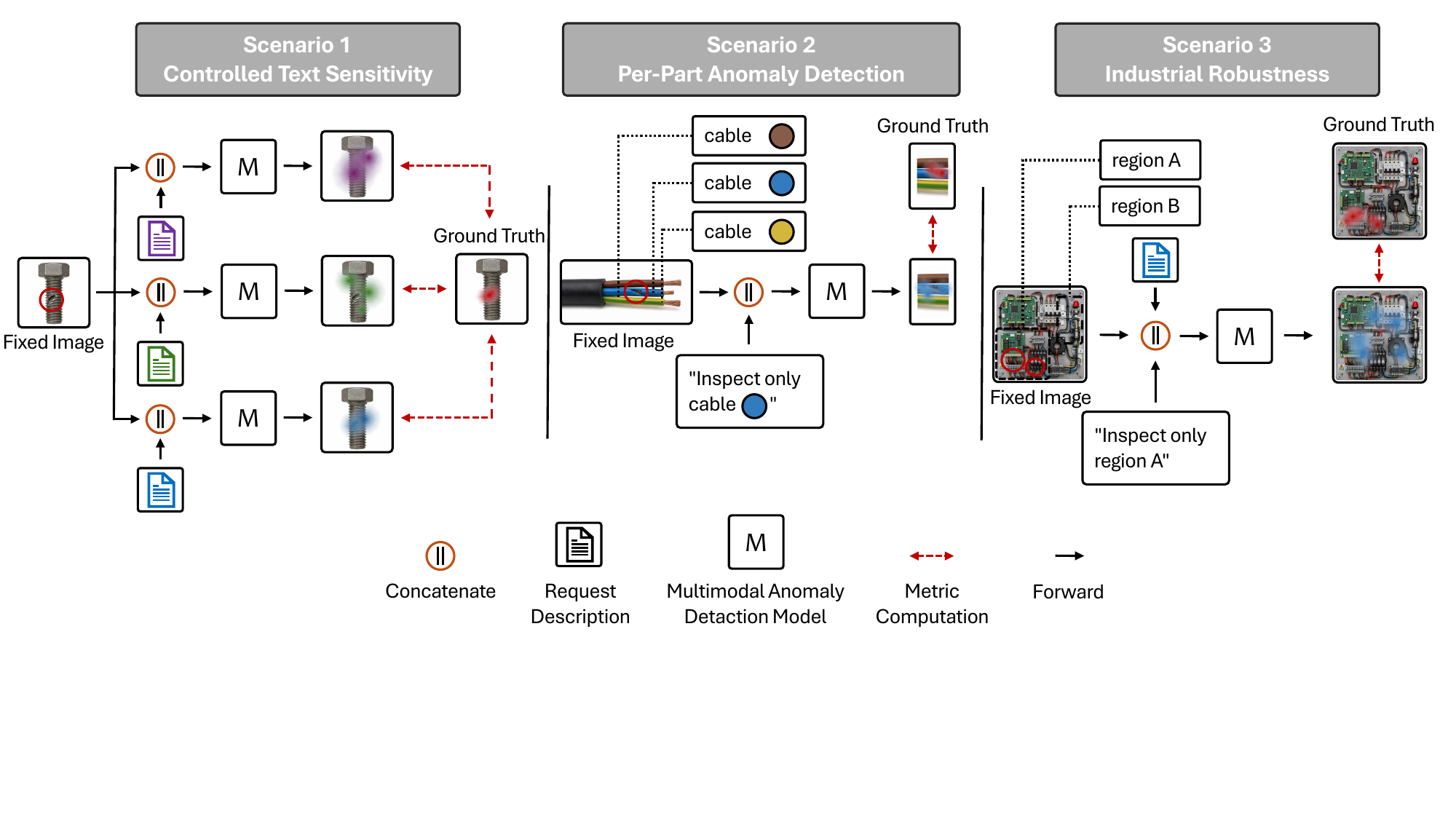}
    \vspace{-75pt}
    \caption{Overview of the three TGAD scenarios, ordered by the functional role language must play in the decision. In each, a fixed image and a textual request are processed by a multimodal anomaly detection model and scored against the ground truth. Scenario~1 varies the prompt on MVTec~AD to test whether the output moves with language; Scenario~2 instructs inspection of a single component, admitting defects in other components as normal; Scenario~3 uses the Assembled Panel dataset, where both the defect type and the component location must be known to inspect.}
    \label{fig:overview}
\end{figure*}

We propose Text-Guided Anomaly Detection (TGAD), a structured benchmark that progressively increases the functional role of language across three scenarios, coupling images with relevant complementary information, summarized in Figure~\ref{fig:overview}. The first, a controlled prompt-sensitivity setting on MVTec~AD, fixes the visual input and ground-truth annotations and varies the textual input systematically, isolating whether predictions move as a function of language at all. The second, a component-tagged extension of MVTec~AD, instructs the model to inspect a specified part of the object; anomalies in other parts are admitted as normal, forcing the model to selectively follow the textual constraint rather than report any visually evident defect. The third, the new Assembled Panel Dataset (APD) made of electronic boards acquired with a domain partner, is the natural convergence of the first two under realistic inspection conditions, where both the defect type and the component location must be known to inspect. The APD and the per-part MVTec~AD extension, together with the per-part validation protocol, are released as part of the benchmark.

To test these claims, we evaluate one representative model from each of the three multimodal anomaly detection paradigms: generative large vision-language reasoning, training-free discriminative pipelines, and embedding-adaptive discriminative methods. Across the three paradigms, the textual interface conditions the decision only superficially. In the first scenario, prompt content is absorbed once an object anchor is present; image-level scores move only when the object/texture noun is removed, a phenomenon we highlight as Object-Anchor Collapse. In the second, component-level instructions fail to constrain the global decision without explicit architectural support, and even with such support, models remain fragile to component overlap and prompt design. Across these stress tests, a consistent diagnostic signature emerges, the Localization-Decision Dissociation: image-level detection collapses while AUPRO localization, which we report alongside the pixel-count-biased pixel-level AUROC \cite{hossain2024evaluating}, degrades comparatively less. On the APD, both limitations converge under realistic load: neither defect-type nor component-location knowledge is accessible through the textual interface.

The contributions of this work are as follows:
\begin{itemize}
\item \textbf{Benchmark and datasets.} TGAD, a structured benchmark for text-guided anomaly detection organized as three scenarios in which language carries a progressively larger share of the inspection decision, the third being the natural convergence of the first two under realistic acquisition conditions. We release the benchmark with a component-tagged extension of MVTec~AD and the APD, a pixel-annotated industrial dataset of electronic panels acquired with domain experts.

\item \textbf{Critical analysis.} A study of one representative model per multimodal anomaly detection paradigm (generative large vision-language, training-free discriminative, embedding-adaptive discriminative), showing that textual input acts as a weak class-level prior rather than an inference mechanism: prompts barely move the decision unless the object noun is removed (the \emph{Object-Anchor Collapse}), and component-level instructions fail to restrict the global decision without explicit architectural support.

\item \textbf{Diagnostic signature and convergence.} The \emph{Localization-Decision Dissociation}, a recurring signature under instruction stress in which image-level detection collapses while region-level localization (AUPRO) degrades far less. On the APD the limitations of the three scenarios appear jointly, with image-level discrimination falling to chance or below despite per-error-type prompting.
\end{itemize}

\section{Related Work}
\label{sec:related_work}

\subsection{Visual Anomaly Detection Paradigms}
\label{sec:related_visual_ad}

Visual anomaly detection (AD) methods are grouped by how normality is represented: reconstruction-based approaches treat the reconstruction error of normal samples as the anomaly signal \cite{an2015variational,sakurada2014anomaly,aebergmann}, with memory-augmented variants preventing strong decoders from reconstructing anomalies \cite{gong2019memorizing}; feature- and embedding-based methods model the distribution of normal features from pretrained encoders via memory banks or statistical estimators \cite{bergmann2020unified,roth2022patchcore,ruff2018deep,samele2022patchwise}; and synthesis-based methods learn boundaries from synthetic anomalies \cite{li2021cutpaste,zavrtanik2021draem,chen2024unified,chen2025progressive}. All three are unimodal, treat any deviation from the learned appearance distribution as anomalous, and cannot incorporate semantic constraints or context-dependent normality \cite{chandola2009anomaly,pang2021deep}. They form the empirical baseline for the multimodal value proposition; our benchmark probes whether the textual interface adds a functional role beyond the visual prior inherited from these backbones.

\subsection{Multimodal Anomaly Detection: Paradigms and Models}
\label{sec:related_multimodal_ad}

Combining vision and language yields more descriptive representations and supports task transfer through linguistic priors rather than per-task training \cite{baltrusaitis2019multimodal}. CLIP \cite{radford2021learning} and large vision-language models brought this paradigm to AD, with WinCLIP \cite{jeong2023winclip} as an early zero/few-shot exemplar. The resulting methods span three architecturally distinct paradigms, for each of which we take one representative: AnomalyGPT~\cite{gu2024anomalygpt} (generative large vision-language), LogSAD~\cite{zhang2025towards} (training-free discriminative), and AA-CLIP~\cite{ma2025aaclip} (embedding-adaptive discriminative); where text enters each pipeline is detailed in \Cref{sec:method_models}. All present the textual interface as enabling instruction-driven inspection, and the natural-language outputs of some are framed as a step beyond anomaly maps toward human-understandable explanations \cite{arrieta2020xai,pang2021deep}. In practice the text consumed at inference is most often a static category-level prompt (a class name, a canonical defect list, a ``flawless object''/``damaged object'' template) rather than a controllable instruction the model must parse and act upon. We treat these models as subjects of critical analysis rather than baselines to outperform, and stress-test whether their textual interface functionally conditions inference---a question their original evaluations do not isolate.

\subsection{Evaluation Approaches for Multimodal Anomaly Detection}
\label{sec:related_eval}

Standard industrial AD evaluation is anchored to MVTec AD \cite{bergmann2021mvtec}, with image-level AUROC, pixel-level AUROC, and the region-weighted AUPRO metric; MVTec LOCO \cite{bergmann2022logical} adds logical anomalies under the same paradigm. Under the class imbalance of industrial images, pixel-level AUROC is dominated by the large non-anomalous pixel population and can stay high even when localization is spatially poor \cite{hossain2024evaluating}, which is why we adopt AUPRO. VisA \cite{zou2022spot} enters our protocol only as the cross-domain adaptation set for one model. A distinct line of work evaluates MLLMs on industrial AD: MMAD \cite{jiang2025mmad} is a VQA benchmark probing whether MLLMs can identify, describe, and localize defects in language; EIAD \cite{zhang2025eiad} generates natural-language explanations; and an MLLM zero-shot AD-and-reasoning framework has also been proposed \cite{xu2025towards}. These measure whether a model can \emph{argue about} defects in words, not whether language \emph{changes} the detection output, and do not produce a continuous anomaly map (hence no pixel-level AUROC or AUPRO on detection). Our benchmark instead binds detection to specific textual conditions and measures their effect on the model's anomaly map directly---distinct from MVTec-style evaluation, which holds text constant, and from MLLM frameworks, which assess linguistic competence about defects rather than the influence of language on the decision.

\section{Methodology}
\label{sec:methodology}

This section defines the TGAD benchmark and its two released datasets, the three evaluated models including the architectural extension applied to LogSAD, and the experimental protocol.
TGAD comprises three scenarios that progressively raise the functional role of language, each built on a dedicated dataset (Figure~\ref{fig:overview}).

\paragraph{Scenario~1: Controlled Text-Sensitivity.} The first scenario isolates whether predictions move as a function of language at all: the visual input, the split, and the ground-truth annotations are fixed while the textual input is systematically varied. By systematic variation we mean a graded sequence of controlled edits that progressively reduce the semantic content of the prompt while the image stays fixed, so that any change in the output can be attributed to the edited component: the object (class) noun, the normality assertion, and the enumerated defect list are removed in turn, down to a minimal content-free query. Each model receives the variant set matching its textual interface, free-form prompts for the generative model and keyword substitutions (a class name, a detailed description, and a generic word) for the keyword-driven discriminative ones; the full templates and per-class defect lists are reported in Appendix~\ref{app:prompts_scen1}. It uses MVTec~AD \cite{bergmann2021mvtec} as-is, with its original train/test split; MVTec~AD contains 15 categories (10 object, 5 texture) with class-specific resolutions up to $1024 \times 1024$, pixel-level annotations on all defective test samples, and defect-free training sets.

\paragraph{Scenario~2: Per-Part Anomaly Detection.} The second scenario instructs the model to inspect a specified component of the object, admitting defects in other components as normal under a controlled validation rule. It is built on a component-tagged extension of MVTec~AD that we release with this paper. For each of nine object classes (the \textit{bottle} class is excluded as a single visual entity without meaningful part decomposition) we annotate every defective test image with the set $T_i \subseteq T$ of component tags whose region is affected by the defect; the tag inventory is class-specific (e.g., for \textit{cable}: light-blue insulation, green cable, blue cable, gray cable). A subclass is defined by a tag set $T_s \subseteq T$ with $|T_s| \in \{1, 2, 3\}$, paired with one of two evaluation settings: under Evaluation Setting~1 (EV1), only defect-free images are admitted as normal; under Evaluation Setting~2 (EV2), defective images whose defects are disjoint from $T_s$ are additionally admitted as normal (irrelevant-defect images), forcing the model to follow the textual instruction rather than respond to any visually evident defect. Formally, for the test set $\mathcal{I}$ of a class and a subclass tag set $T_s$:
\begin{align}
\mathcal{A}(T_s)   &= \{\,i \in \mathcal{I} : T_i = T_s\,\}, \nonumber \\
\mathcal{N}_1(T_s) &= \{\,i \in \mathcal{I} : T_i = \emptyset\,\}, \nonumber \\
\mathcal{N}_2(T_s) &= \{\,i \in \mathcal{I} : T_i \neq \emptyset,\; T_i \cap T_s = \emptyset\,\}
\label{eq:subclass}
\end{align}
where $T_i$ is the set of component tags whose region is affected by a defect in image $i$ ($T_i = \emptyset$ denotes a defect-free image). The normal set is $\mathcal{N}_1$ under EV1 and $\mathcal{N}_1 \cup \mathcal{N}_2$ under EV2; images with $T_i \cap T_s \neq \emptyset$ and $T_i \neq T_s$ overlap the instructed part only partially and are excluded under both settings, removing label ambiguity. The full tag inventory, the subclass mapping for all classes, per-subclass cardinalities, the two evaluation settings, and the LLM-randomized per-subclass prompt pool used at inference are released with the benchmark and detailed in Appendix~\ref{app:perpart}. Image-level metrics use $\mathcal{A}(T_s)$ against the normal set of the chosen setting; localization metrics (AUPRO, P-AUROC) are computed on $\mathcal{A}(T_s)$ with the original MVTec ground-truth masks. EV2 therefore alters only the image-level negative set, while localization is scored on the same in-scope defects under both settings, which is why a model can keep AUPRO high while its image-level decision collapses.

\begin{figure*}[t]
    \centering
    \includegraphics[width=\linewidth]{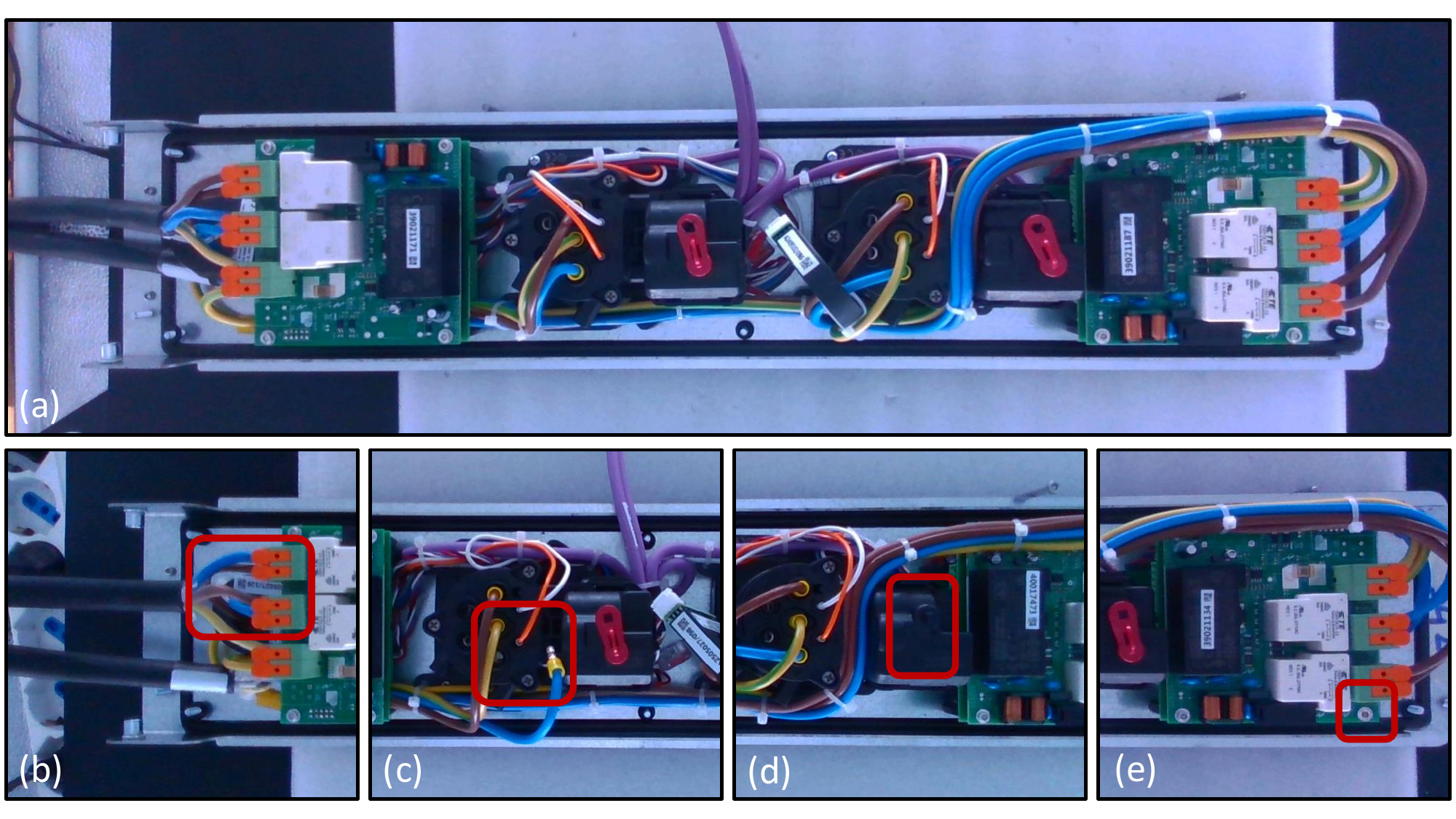}
    \caption{Example of the Assembled Panel Dataset. (a)~overall panel layout; (b)--(e)~close-ups of representative defects: swapped power-cable colors at a power port, an unplugged wire at an EV port, a missing red protective cap, and a missing screw on a PCB.}
    \label{fig:apd_example}
\end{figure*}

\paragraph{Scenario~3: Industrial Robustness.} The third scenario is the natural convergence of the first two under realistic inspection conditions, where defect-type and component-location knowledge are simultaneously necessary to inspect. It is built on the new APD, released with this paper and acquired with an industrial partner manufacturing assembled electrical panels (Figure~\ref{fig:apd_example}): each panel holds multiple interacting components (printed circuit boards, connectors, sensors, cables, labels, screws) in a fixed layout, imaged in a fixed photo-box setup so that visual variability across acquisitions reflects anomalies rather than scene factors. The taxonomy, defined with domain experts, comprises 14 categories grouped by reversibility into three severity types, from non-reversible damage to minor reversible defects (for example, swapped power-cable colors, missing components, or incorrect wire routing). The dataset contains 99 defect-free and 53 anomalous images at $1920 \times 1080$, pixel-level annotated, with some samples carrying more than one defect category. The full specification (industrial context, acquisition, taxonomy with per-category counts, defect creation, and annotation protocol) is provided in Appendix~\ref{app:assembled_panel}.

We evaluate the APD in two modes. We first measure performance in a general full-image setup (\textit{Standard AD}), using a single prompt shared across all categories with several levels of normal object descriptions. We then introduce the structured text-guided prompts (\textit{Per-Error AD}): for each error type, a dedicated prompt instructs the model to focus on the affected component, exploiting per-part anomaly detection in images of high visual complexity. As in Scenario~2, the prompt set matches each model's textual interface; all templates and information are reported in Appendices~\ref{app:assembled_panel} and~\ref{app:prompts:scenario3_std}. Only defect-free images serve as normal samples, since the scenario is challenging on its own. In the Per-Error AD scenario, we evaluate each defect-type set of images against all the normal one, and then average the results.

\subsection{Evaluated Models}
\label{sec:method_models}

We evaluate AnomalyGPT \cite{gu2024anomalygpt} as the generative large vision-language paradigm, LogSAD \cite{zhang2025towards} as the training-free discriminative paradigm, and AA-CLIP \cite{ma2025aaclip} as the embedding-adaptive discriminative paradigm. Generic Vision-Language Models (VLMs) are excluded from our investigation, as they do not intrinsically produce anomaly maps, which are essential for measuring localization capabilities. The paragraphs below state only where textual input enters each pipeline and the adaptations applied to make the three models comparable across the benchmark; full architectural details are deferred to the cited papers.

\paragraph{AnomalyGPT.} We evaluate AnomalyGPT in both zero-shot and 1-shot regimes. The dual scoring mechanism (zero-shot patch-to-text, few-shot patch-to-patch) is the critical point for our analysis: textual input affects the anomaly map only in zero-shot, while few-shot scoring reduces to visual reference matching.

\paragraph{LogSAD.} We evaluate LogSAD in 1-shot and full-shot regimes. As released, LogSAD does not support per-part inspection: foreground localization with category-level keywords selects the whole object, so component-level prompts do not enforce spatial focus. We therefore introduce the extension described next.

As released, LogSAD treats the entire object as foreground because CLIP compares all patches to a single global text embedding. In general, pure visual models have a documented limitation for objects with multiple visually related parts \cite{Samele2026SADSeM}; component-level prompts therefore do not enforce spatial focus. To enable per-part inspection, we introduce an explicit foreground-masking mechanism. Given a focus prompt encoding the target component and a set $\mathcal{K}_{\text{contrast}}$ of contrast prompts encoding the remaining components, with corresponding CLIP text embeddings $\mathbf{t}_{\text{focus}}$ and $\{\mathbf{t}_{\text{contrast}_k}\}_{k \in \mathcal{K}_{\text{contrast}}}$, we define for each visual patch $j$ with embedding $\mathbf{p}_j$:
\begin{align}
s^{\text{focus}}_j   &= \cos\!\bigl(\mathbf{p}_j,\, \mathbf{t}_{\text{focus}}\bigr), \nonumber \\
s^{\text{contrast}}_j &= \max_{k \in \mathcal{K}_{\text{contrast}}}\, \cos\!\bigl(\mathbf{p}_j,\, \mathbf{t}_{\text{contrast}_k}\bigr), \nonumber \\
\mathrm{mask}_j      &= \mathbbm{1}\!\left[\, s^{\text{focus}}_j > s^{\text{contrast}}_j \,\right]
\label{eq:fg_mask}
\end{align}
where $\cos(\cdot,\cdot)$ is the cosine similarity, $\mathbbm{1}[\cdot]$ is the indicator function, and the resulting per-patch indicators are collected into a binary foreground mask $\mathbf{M}_{\text{fg}} \in \{0,1\}^N$ over the $N$ patches of the input. The mask is then applied element-wise to the structural anomaly map produced by LogSAD's standard pipeline:
\begin{equation}
\mathbf{A}_{\text{masked}} = \mathbf{A}_{\text{structural}} \odot \mathbf{M}_{\text{fg}}
\label{eq:masked_map}
\end{equation}
where $\mathbf{A}_{\text{structural}} \in \mathbb{R}^N$ is the per-patch anomaly map from LogSAD and $\odot$ denotes element-wise multiplication; $\mathbf{A}_{\text{masked}}$ replaces $\mathbf{A}_{\text{structural}}$ for per-part scoring. The extension is applied in Scenario~2 (per-part on MVTec AD) and where appropriate in Scenario~3 (per-error-type prompts on Assembled Panel); for Scenario~1 the unmodified LogSAD pipeline is used.

\paragraph{AA-CLIP.} We trained AA-CLIP on VisA~\cite{zou2022spot}, a dataset disjoint from both MVTec AD and Assembled Panel, preserving the zero-shot character with respect to evaluation. The choice is motivated by the superior performance obtained with the VisA-based training strategy shown in the original paper.
The default CLIP backbone operates at $336 \times 336$; we run a separate adaptation pass at $224 \times 224$ to enable cross-model comparison, since positional embeddings and early-layer adapters become resolution-dependent after adaptation.

\subsection{Experimental Protocol}
\label{sec:method_protocol}

\paragraph{Metrics.} We report image-level AUROC (I-AUROC) and pixel-level AUROC (P-AUROC) as standard threshold-independent measures of global discrimination and per-pixel localization. Under the heavy class imbalance characteristic of industrial AD, P-AUROC is dominated by the large non-anomalous pixel population and can remain high even when the spatial precision of the anomaly map is poor~\cite{hossain2024evaluating}; we therefore commit to AUPRO~\cite{bergmann2022logical} as the localization metric of record. AUPRO is the area under the PRO-FPR curve with FPR computed on normal pixels only, where PRO is the underlying Per-Region-Overlap metric; because each ground-truth region contributes equally regardless of pixel extent, AUPRO neutralizes the imbalance bias inherent to per-pixel scoring.

\paragraph{Supervision Regimes.} The benchmark considers three supervision regimes: zero-shot (no reference defect-free images at inference); 1-shot (one reference defect-free image for visual similarity matching); and full-shot (all defect-free training images, applicable to LogSAD's coreset construction). Each model is evaluated in the regimes for which it is designed: AnomalyGPT in zero-shot and 1-shot under its original protocol; LogSAD in 1-shot and full-shot; AA-CLIP in zero-shot after VisA adaptation.

\paragraph{Visual Input Handling.} Because the models operate in a square aspect-ratio setting and the APD consists of rectangular images, we apply two standard pre-processing strategies, used for all three models, that control the effective visual content while respecting the input-size constraint. \textit{Direct Resize} uniformly resizes the image to the target square resolution. \textit{Pad to Square} first resizes the image to a smaller effective resolution and then pads it with zero-valued pixels to the model's expected input size.

\section{Experiments and Results}
\label{sec:experiments}

The three subsections below report results for Scenarios 1, 2, and 3 in order, each in the same cross-model order. All experiments ran on a single NVIDIA Quadro RTX 6000 (24\,GB). Per-class breakdowns and full per-resolution sweeps are deferred to the appendix.

\subsection{Scenario 1: Controlled Text-Sensitivity}
\label{sec:res_scen1}

\begin{table*}[t]
\centering
\caption{Effect of textual prompt variants on MVTec~AD; all-class average at $224\times224$.}
\label{tab:prompt_ablation_all_models}
\small
\begin{tabular*}{\textwidth}{@{\extracolsep{\fill}}llccc@{}}
\hline
\textbf{Model} & \textbf{Prompt variant} & \textbf{I-AUROC} & \textbf{P-AUROC} & \textbf{AUPRO} \\
\hline
AnomalyGPT
& Original Description     & 97.4 & 93.1 & 90.0 \\
(0-shot) & New Description          & 97.4 & 93.1 & 90.0 \\
& No Anomaly Description   & 97.4 & 93.1 & 90.0 \\
& No Normality Description & 97.4 & 93.1 & 90.0 \\
& Minimal Prompt           & 82.6 & 89.9 & 80.1 \\
\hline
AnomalyGPT
& Original Description     & 94.2 & 95.5 & 89.5 \\
(1-shot) & Minimal Prompt          & 94.2 & 95.5 & 89.5 \\
\hline
LogSAD
& Class name & 89.1 & 95.4 & 85.8 \\
(1-shot) & Detailed   & 89.1 & 95.4 & 85.8 \\
& Generic    & 89.1 & 95.4 & 85.8 \\
\hline
LogSAD
& Class name & 98.0 & 97.8 & 92.3 \\
(full-shot) & Detailed   & 98.0 & 97.8 & 92.3 \\
& Generic    & 98.0 & 97.8 & 92.3 \\
\hline
AA-CLIP
& Class name & 84.5 & 90.9 & 81.1 \\
(0-shot) & Detailed   & 84.6 & 90.9 & 81.3 \\
& Generic    & 84.9 & 91.0 & 81.4 \\
\hline
\end{tabular*}
\end{table*}

\paragraph{AnomalyGPT.} The AnomalyGPT zero-shot baseline reaches $97.4/93.1/90.0$ (I-AUROC/P-AUROC/AUPRO), reproducing the original paper within precision and providing the reference for prompt invariance (Table~\ref{tab:prompt_ablation_all_models}). Of the five variants, the first four (Original, paraphrase, defect list removed, normality assertion removed) are identical to the reported precision; only the Minimal prompt, which additionally removes the object noun, moves the scores, and it moves all three metrics together ($97.4 \to 82.6$ I-AUROC, $93.1 \to 89.9$ P-AUROC, $90.0 \to 80.1$ AUPRO). The model is anchored on the object noun alone: paraphrase, defect list, and normality assertion are inert, and only deleting the noun triggers the Object-Anchor Collapse. Image-level I-AUROC drops more than the pixel-count-biased P-AUROC, the first appearance of the Localization-Decision Dissociation. We also report the 1-shot regime as a control: there AnomalyGPT scores each patch by visual similarity to the reference image, so the prompt sits outside the scoring path and, as expected, leaves the scores unchanged ($94.2/95.5/89.5$ for both Original and Minimal). The zero-shot Object-Anchor Collapse is therefore a property of the text path specifically, not a generic insensitivity of the model.

\paragraph{LogSAD.} LogSAD's text path enters only at the foreground stage; we test three keyword forms (exact class name, semantic synonym, generic ``object''/``texture''). All three yield identical scores in both the 1-shot ($89.1/95.4/85.8$) and full-shot ($98.0/97.8/92.3$) regimes, because LogSAD selects the entire MVTec image as foreground under any keyword: no MVTec image has a meaningful background to suppress, so keyword choice never alters the foreground region. The textual interface is therefore functionally disconnected from the decision, and the large full-shot over 1-shot gain shows that performance is driven by the visual matching pathway rather than by language.

\paragraph{AA-CLIP.} AA-CLIP shows the only non-zero keyword sensitivity, but it is negligible: at $224 \times 224$ all-class I-AUROC moves between $84.5$ and $84.9$ across the original, class-name, and generic keywords, with P-AUROC and AUPRO essentially flat ($\approx 91$ and $\approx 81$). The generic word marginally outscores the class-specific keyword ($84.9$ vs $84.6$), already indicating that the keyword carries no useful conditioning.

\paragraph{} Across the three models no prompt edit changes the operating regime. The single largest movement is AnomalyGPT's Object-Anchor Collapse, triggered only by removing the object noun; LogSAD is exactly invariant; AA-CLIP moves by less than half a point. In every case the textual input behaves as a static class prior rather than a controllable instruction, and where image-level scores do move they move more than the localization metrics, the recurring Localization-Decision Dissociation. The full prompt sets for all three models are listed in Appendix~\ref{app:prompts_scen1}.

\subsection{Scenario 2: Per-Part Anomaly Detection}
\label{sec:res_scen2}

\begin{table*}[t]
\centering
\caption{Per-part anomaly detection on the MVTec~AD extension (all-class average, $224\times224$), under EV1 and EV2. Under EV2 (defects outside the instructed part admitted as normal) image-level scores drop sharply while AUPRO holds; the foreground-masking extension (FMS, Equations~\ref{eq:fg_mask}--\ref{eq:masked_map}) narrows the EV1$\to$EV2 gap at a localization cost. LogSAD is reported as released and with FMS.}
\label{tab:perpart_all_models}
\footnotesize
\begin{tabular*}{\textwidth}{@{\extracolsep{\fill}}llcccccc@{}}
\hline
\textbf{Model} & \textbf{\# Parts}
& \multicolumn{3}{c}{\textbf{EV1}}
& \multicolumn{3}{c}{\textbf{EV2}} \\
\cmidrule(lr){3-5}\cmidrule(lr){6-8}
&
& \textbf{I-AUROC} & \textbf{P-AUROC} & \textbf{AUPRO}
& \textbf{I-AUROC} & \textbf{P-AUROC} & \textbf{AUPRO} \\
\hline
AnomalyGPT
& 1 part  & 90.3 & 95.8 & 89.2 & 66.3 & 94.4 & 84.9 \\
(0-shot) & 2 parts & 94.2 & 93.7 & 85.8 & 66.6 & 92.0 & 80.0 \\
& 3 parts & 99.1 & 97.9 & 92.5 & 80.5 & 96.3 & 86.5 \\
\hline
AA-CLIP
& 1 part  & 79.1 & 93.6 & 81.5 & 64.7 & 93.1 & 79.9 \\
(0-shot) & 2 parts & 78.3 & 87.8 & 76.1 & 69.2 & 87.6 & 75.4 \\
& 3 parts & 77.5 & 86.0 & 75.2 & 75.9 & 85.8 & 75.1 \\
\hline
LogSAD
& 1 part  & 97.6 & 98.7 & 95.5 & 73.0 & 97.7 & 92.2 \\
(full-shot) & 2 parts & 97.9 & 97.8 & 92.3 & 83.7 & 97.4 & 91.0 \\
& 3 parts & 96.3 & 98.4 & 90.2 & 91.8 & 98.5 & 90.4 \\
\hline
LogSAD
& 1 part  & 73.6 & 69.3 & 52.8 & 65.1 & 69.1 & 52.1 \\
(full-shot, FMS) & 2 parts & 89.9 & 85.6 & 76.1 & 80.2 & 85.4 & 75.5 \\
& 3 parts & 99.5 & 98.0 & 91.7 & 95.0 & 98.0 & 91.9 \\
\hline
\end{tabular*}
\end{table*}

Scenario~2 stress-tests whether a model restricts anomaly assessment to the instructed component by admitting defective images with unrelated defects as normal. The EV1\,$\to$\,EV2 transition is the diagnostic: under EV1 a model that flags any defect is correct, whereas under EV2 it must additionally suppress responses to defects outside the instructed component. Table~\ref{tab:perpart_all_models} reports the all-class results for the three models, with LogSAD in its released form and with the foreground-masking extension (FMS).

\paragraph{AnomalyGPT.} Under EV1, AnomalyGPT performs well at one part (I-AUROC $90.3$, AUPRO $89.2$) and improves toward standard AD as more parts are admitted (three parts: $99.1$, $92.5$). Under EV2 the image-level score collapses while localization barely moves: at one part I-AUROC drops $90.3 \to 66.3$, against $89.2 \to 84.9$ in AUPRO and $95.8 \to 94.4$ in P-AUROC. The pattern holds at two parts ($66.6$ I-AUROC, $80.0$ AUPRO) and weakens at three ($80.5$, $86.5$), where admitting three components brings the task close to standard AD. The dissociation is therefore sharpest at one part, where the instruction is most restrictive: AnomalyGPT localizes present defects correctly but cannot suppress its image-level response to defects in non-instructed components.

\paragraph{LogSAD.} As released, LogSAD is the strongest model under EV1 (I-AUROC $97.6$/$97.9$/$96.3$) yet shows the same dissociation under EV2: the one-part transition is $97.6 \to 73.0$ in I-AUROC (a $24.6$-point drop) against $95.5 \to 92.2$ in AUPRO, and the gap narrows as more parts are admitted ($97.9 \to 83.7$ at two parts, $96.3 \to 91.8$ at three). The foreground-masking extension (FMS, Equations~\ref{eq:fg_mask}--\ref{eq:masked_map}) is the only mechanism that narrows this gap: at one part it cuts the EV1\,$\to$\,EV2 I-AUROC drop from $24.6$ to $8.5$ points ($73.6 \to 65.1$), confirming that explicit spatial conditioning recovers instruction-following. The recovery is paid in localization, with EV1 one-part AUPRO falling from $95.5$ to $52.8$ as the mask over-restricts or excludes anomalous regions. At three parts, where most of the object is admitted as foreground, FMS instead improves both I-AUROC ($96.3 \to 99.5$) and AUPRO ($90.2 \to 91.7$).

\paragraph{LogSAD masking: qualitative analysis.} The extension behaves well when components are spatially separable and the focus region preserves its semantic appearance under defect, as in the \textit{hazelnut} class: the mask isolates the target component and the anomaly inside it is detected. Three failure modes characterize the masking fragility: (i) spatial overlap of components prevents clean isolation (the \textit{capsule} class, where the white printed ``500'' lies on top of the orange half); (ii) defects that alter the focus region's visual appearance can move it below the contrast threshold and exclude the anomalous patches (a crack on the \textit{capsule} black half that recolors it toward red); (iii) structured-background regions can be semantically close to the focus and leak into the foreground mask (the \textit{transistor} class background). Each mode follows from the foreground stage relying on cosine similarity in a shared CLIP space without explicit spatial regularization.

\begin{figure*}[t]
    \centering
    \includegraphics[width=\linewidth]{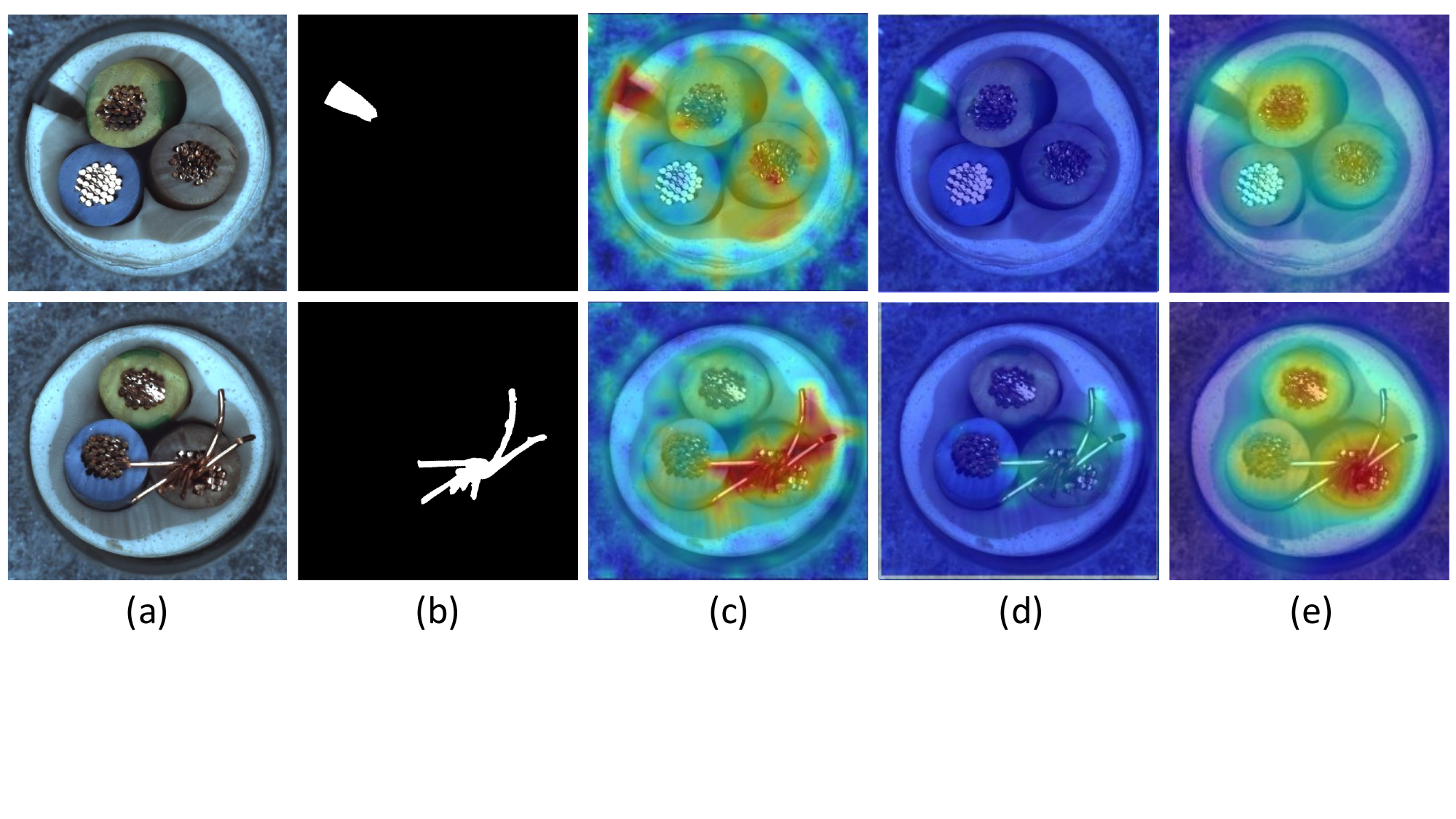}
    \caption{Qualitative cross-model comparison on the \textit{cable} class under EV2, for a one-part request to inspect only the grey sub-cable. Columns: input image (a), ground truth (b), and the anomaly maps of AnomalyGPT (c), LogSAD (d), and AA-CLIP (e). Top row: a defect on the non-instructed outer insulation, to which every model responds; bottom row: a defect on the instructed grey wire, which all models localize correctly. The component-level instruction fails to suppress the response to the unrelated defect.}
    \label{fig:perpart_mask_fail}
\end{figure*}

\paragraph{AA-CLIP.} AA-CLIP shows the dissociation on a markedly lower baseline, and, unlike the other two models, its EV1 I-AUROC does not improve as more parts are admitted but slightly declines ($79.1/78.3/77.5$): an early sign that its image-level score reads a pooled global statistic rather than the instructed components, the mechanism that collapses in Scenario~3. Under EV2 the one-part transition is $79.1 \to 64.7$ (AUPRO $81.5 \to 79.9$), shrinking at two parts ($78.3 \to 69.2$) and almost vanishing at three ($77.5 \to 75.9$) as the task approaches standard AD. Replacing class-level with component-specific keywords does not enforce spatial focus, consistent with the documented limitation of CLIP global text-patch similarity; lacking a masking analogue, AA-CLIP cannot recover instruction-conditioning here.

\paragraph{} Across all three models the EV2 dissociation is the same: image-level discrimination collapses, AUPRO degrades comparatively less, and the pixel-count-biased P-AUROC barely moves (at one part $95.8 \to 94.4$ for AnomalyGPT and $98.7 \to 97.7$ for LogSAD), which confirms AUPRO as the informative localization metric. Figure~\ref{fig:perpart_mask_fail} illustrates failures on the \textit{cable} class under a one-part EV2 request to inspect the grey sub-cable: every model (columns c--e) responds to a defect on the non-instructed outer insulation (top row) while correctly localizing a defect on the instructed wire (bottom row). LogSAD with FMS is the only configuration that measurably restores instruction-conditioning, at an AUPRO cost on EV1 and with the three qualitative failure modes above; component-level control is therefore not an emergent property of the evaluated models. The full per-subclass prompt sets are listed in Appendix~\ref{app:prompts_scen2}.

\subsection{Scenario 3: Industrial Robustness on Assembled Panel}
\label{sec:res_scen3}

\begin{table}[t]
\centering
\footnotesize
\setlength{\tabcolsep}{3pt}
\caption{Assembled Panel at $224\times224$: best configuration per model under Standard~AD and Per-Error~AD. Per-Error prompting raises AUPRO for AnomalyGPT only and lifts the decision for no model.}
\label{tab:model_config_results}
\begin{tabular}{ll ccc}
\toprule
Model & Mode & I-AUROC & P-AUROC & AUPRO\\
\midrule
\multirow{2}{*}{AnomalyGPT} & Standard  & 78.6 & 81.6 & 22.8\\
                            & Per-Error & 71.2 & 85.0 & 54.9\\
\multirow{2}{*}{LogSAD}     & Standard  & 62.6 & 73.1 & 44.5\\
                            & Per-Error & 50.5 & 65.5 & 41.3\\
\multirow{2}{*}{AA-CLIP}    & Standard  & 33.2 & 84.9 & 56.6\\
                            & Per-Error & 31.5 & 84.2 & 51.2\\
\bottomrule
\end{tabular}
\end{table}

\begin{figure}[t]
\centering
\includegraphics[width=\linewidth]{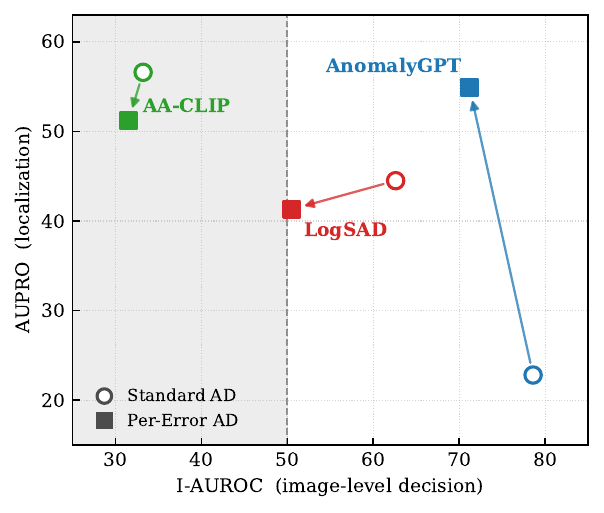}
\caption{Assembled Panel in the I-AUROC--AUPRO plane (best configuration per model); arrows run from Standard~AD to Per-Error~AD. The shaded band marks the level of a random classifier.}
\label{fig:panel_scatter}
\end{figure}

\begin{figure*}[t]
    \centering
    \includegraphics[width=\linewidth]{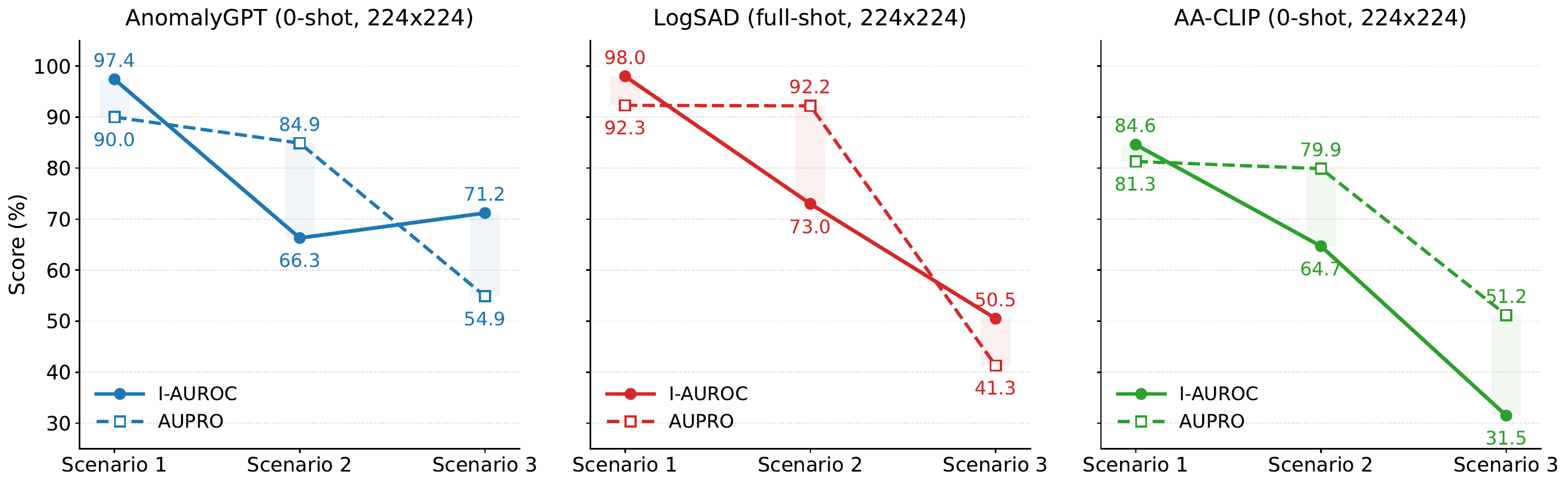}
    \caption{Per-model I-AUROC (solid) and AUPRO (dashed) across the three scenarios; the shaded band is the gap between them. Under the component instruction of Scenario~2 the decision drops below localization for all three models (the Localization-Decision Dissociation); on Assembled Panel localization stays above the decision only for AA-CLIP. Scenario~3 uses Per-Error~AD (Table~\ref{tab:model_config_results}); all at $224\times224$, AnomalyGPT and AA-CLIP zero-shot, LogSAD full-shot.}
    \label{fig:synthesis}
\end{figure*}

Assembled Panel realizes the Scenario~1 and Scenario~2 requirements at once: the panel holds numerous interacting components whose anomalies are subtle deviations in presence, position, or connection rather than localized corrupted patterns, so the model must know what defect to look for and where.
We propose two modes at $224\times224$ resolution, \textit{Standard~AD} and \textit{Per-Error~AD}. Standard AD proposes a single prompt for all images, always declined at different levels of semantic complexity. Per-Error AD queries the model about the presence of a specific defect. Both modes are evaluated against normal images only, but Per-Error AD evaluates each defect type alone and then averages across all defect types.
Table~\ref{tab:model_config_results} gives the best configuration (we vary the visual input handling mode) per model, and Figure~\ref{fig:panel_scatter} places every (model, mode) in the I-AUROC--AUPRO plane; the full per-prompt and per-resize sweeps are deferred to Appendix~\ref{app:ap:extended}, which also shows that no single input handling is best across models---AnomalyGPT prefers Direct Resize, LogSAD Pad to Square---so the per-model ``best configuration'' is itself a deployment caveat.

\paragraph{Image-level detection collapses for all three.} No model approaches its MVTec level. AnomalyGPT is the strongest on the decision (Standard~AD I-AUROC $78.6$, zero-shot) but localizes poorly (AUPRO $22.8$), LogSAD reaches $62.6$, and AA-CLIP drops to $33.2$. Neither extra text nor a reference image helps: prompt richness hurts rather than aids AnomalyGPT (its verbose prompts score $71.7$ against $78.6$ for the minimal one), the prompt is inert for LogSAD since the whole panel is selected as foreground exactly as in Scenario~1, and AnomalyGPT's few-shot regime underperforms its zero-shot one ($1$-shot I-AUROC $65.6$, AUPRO $17.8$, Direct Resize), the visual-matching pathway itself breaking down on the cluttered layout.

\paragraph{Text steers localization, not the decision.} Per-Error~AD raises AnomalyGPT's AUPRO sharply ($22.8\to54.9$) while leaving its I-AUROC essentially unchanged ($78.6\to71.2$): naming the component guides \emph{where} the model looks but does not condition the decision. For LogSAD the foreground-masking extension does not recover performance here ($50.5$ I-AUROC, $41.3$ AUPRO); as in Scenario~2 it trades image-level discrimination for instruction-following, and the recovery is weaker than on MVTec because the components are smaller, more numerous, and visually closer. AA-CLIP responds to neither mode.

\paragraph{AA-CLIP: a structural pooling effect.} The weakness foreshadowed in Scenario~2 becomes the dominant failure here. AA-CLIP's image-level discrimination is structurally non-functional on the panel, staying in the $31$--$33$ range while its AUPRO is the highest of the three ($51$--$57$) and P-AUROC remains high ($84$--$85$). The cause is the same global pooling: aggregating patch--text similarities into a single image embedding, so on a high-resolution scene with sparse component-level anomalies the large normal-patch population dilutes the anomaly signal at aggregation, while the patch-level signal that AUPRO measures stays informative. In Figure~\ref{fig:panel_scatter} this isolates AA-CLIP in the low-I-AUROC, high-AUPRO corner.

\paragraph{} Under realistic load the picture is harsher than a simple dissociation. Image-level discrimination is uniformly low, but localization no longer holds across the board: relative to MVTec, AUPRO degrades as much as I-AUROC for the matching-based AnomalyGPT and LogSAD, whose region-level signal erodes on the cluttered panel; the Localization-Decision Dissociation survives only for AA-CLIP. Per-Error prompting lifts localization for AnomalyGPT alone and the decision for no model. The realistic setting therefore reproduces the Scenario~1 and Scenario~2 limitations and adds one: for the matching paradigms, localization itself breaks down.

\subsection{Synthesis Across Scenarios}
\label{sec:res_final}

Each paradigm fails through one mechanism that carries from the controlled to the realistic setting. In Scenario~1 the textual interface is absorbed (object-anchor in AnomalyGPT, whole-image foreground in LogSAD, negligible keyword sensitivity in AA-CLIP). In Scenario~2 component-level instructions do not constrain the global decision without architectural support---LogSAD's foreground masking is the only mechanism that recovers conditioning, at an AUPRO cost, while AA-CLIP signals its weakness by failing to improve as the task is eased---and the Localization-Decision Dissociation is here sharpest and universal: the decision collapses while region-level localization holds, for all three models. On Assembled Panel the limitations converge but the dissociation does not survive uniformly: localization erodes alongside the decision for AnomalyGPT and LogSAD, and holds only for AA-CLIP, whose global pooling keeps AUPRO informative while the decision falls below chance. The constant across every setting, hidden by the pixel-count-biased P-AUROC, is that text \emph{selects but does not instruct}. Figure~\ref{fig:synthesis} traces per-model I-AUROC and AUPRO across the scenarios.

\section{Conclusions}
\label{sec:conclusions}

We introduced Text-Guided Anomaly Detection (TGAD), a benchmark whose three scenarios ask language to carry a progressively larger share of the inspection decision, together with two released datasets: a component-tagged extension of MVTec~AD and the Assembled Panel Dataset. Across one representative model per multimodal paradigm (AnomalyGPT, LogSAD, AA-CLIP), the textual interface conditions the decision only superficially---prompts are absorbed once an object anchor is present (Object-Anchor Collapse), component-level instructions do not restrict the global decision without architectural support, and where language leaves a trace it moves localization rather than the decision (the Localization-Decision Dissociation), a pattern that converges under the realistic load of Assembled Panel. In short, these detectors use language as a \emph{prior}, not a \emph{control signal}: the prompt selects what the model already finds, but cannot tell it what to ignore.

This is not an inherent limit---our foreground-masking extension restores instruction-following on LogSAD, at a localization cost---but a consequence of pipelines in which the image-level decision never sees the instruction. The open question is therefore not whether a model can \emph{describe} a defect, which recent work answers well, but whether the instruction can \emph{change} the decision, which TGAD is built to measure. It also points to how the gap might close: the EV2 setting is itself a supervision signal for instruction-following, turning the benchmark into a training curriculum; beyond it lie scoring architectures that condition the decision---not only the anomaly map---on the instruction, and a move from naming an object to specifying a correct configuration, the relational structure that visual priors cannot encode. The aim throughout is to make language a control signal on the decision rather than a prior on it.


\bibliographystyle{plainnat}
\bibliography{egbib}

\newpage
\onecolumn
\appendix


\section{Per-Part MVTec~AD Extension: Full Annotation Specification}
\label{app:perpart}

This appendix specifies the artifacts required to reproduce the per-part scenario from MVTec~AD: the class-wise tag inventory, the subclass-to-tag mapping, the per-subclass cardinalities of the anomalous and normal sets under EV1 and EV2, and the annotation protocol. The per-subclass prompt pool is reported, by model, in Appendix~\ref{app:prompts_scen2}.

\subsection{Tag inventory per class}
\label{app:perpart:tags}

The extension covers nine of the fifteen MVTec~AD classes; six classes (carpet, grid, leather, tile, wood, bottle) are excluded because the first five are texture classes without a meaningful component decomposition and \emph{bottle} is a single visual entity without distinct parts. For each annotated class, Table~\ref{tab:perpart_tags} lists the component tags with one-line definitions.

\begin{table}[h]
\centering
\caption{Component-tag inventory for the nine annotated MVTec~AD classes; each tag names a visually distinct part of the object.}
\label{tab:perpart_tags}
\small
\begin{tabular}{lp{0.74\textwidth}}
\hline
\textbf{Class} & \textbf{Component tags description} \\
\hline
cable        & light-blue insulation layer (cable1), green cable (cable2), blue cable (cable3), gray cable (cable4) \\
capsule      & white printed ``500'' (capsule1), left black part (capsule2), right reddish-orange part (capsule3) \\
hazelnut     & shell (hazelnut1), hilum (hazelnut2) \\
metal nut   & outer hexagonal body (metal nut1), central threaded hole (metal nut2) \\
pill         & plain surface (pill1), embossed letters ``FF'' (pill2)\\
screw        & threaded shaft (screw1), screw head (screw2), neck (screw3) \\
toothbrush   & plastic base (toothbrush1), white bristles (toothbrush2), colored bristles (toothbrush3)\\
transistor   & plastic body (transistor1), metallic legs (transistor2)\\
zipper       & interior fabric (zipper1), zipper teeth (zipper2), fabric at the edge (zipper3) \\
\hline
\end{tabular}
\end{table}

\subsection{Per-class subclass mapping}
\label{app:perpart:subclasses}

A subclass is a tag set $T_s \subseteq T$ with $|T_s| \in \{1, 2, 3\}$. The admitted subclasses per class are determined by the rule: a tag combination becomes a subclass if and only if at least one defective image in the test split has that exact tag set. Table~\ref{tab:perpart_subclasses} reports the admitted subclasses per class, grouped by cardinality. Subclasses are denoted by concatenating tag indices (e.g., $\{$cable1, cable3$\}$ is written \texttt{cable1\_cable3}).

\begin{table}[h]
\centering
\caption{Admitted subclasses per class, grouped by tag-set cardinality.}
\label{tab:perpart_subclasses}
\small
\begin{tabular}{llll}
\hline
\textbf{Class} & \textbf{1-tag} & \textbf{2-tag} & \textbf{3-tag} \\
\hline
cable      & cable1, cable2,                & cable1\_cable2, cable1\_cable3,     & cable2\_cable3\_cable4 \\
           & cable3, cable4                 & cable2\_cable3, cable2\_cable4,     & \\
           &                                & cable3\_cable4                      & \\
capsule    & capsule1, capsule2,            & capsule1\_capsule3,                 & --- \\
           & capsule3                       & capsule2\_capsule3                  & \\
hazelnut   & hazelnut1, hazelnut2           & hazelnut1\_hazelnut2                & --- \\
metal\,nut & metal\_nut1, metal\_nut2       & metal\_nut1\_metal\_nut2            & --- \\
pill       & pill1, pill2                   & pill1\_pill2                        & --- \\
screw      & screw1, screw2, screw3         & ---                                 & --- \\
toothbrush & toothbrush1,                   & toothbrush1\_toothbrush2,           & toothbrush1\_toothbrush2\_ \\
           & toothbrush2,                   & toothbrush2\_toothbrush3            & toothbrush3 \\
           & toothbrush3                    &                                     & \\
transistor & transistor1, transistor2       & transistor1\_transistor2            & --- \\
zipper     & zipper1, zipper2,              & zipper1\_zipper2,                   & --- \\
           & zipper3                        & zipper1\_zipper3,                   & \\
           &                                & zipper2\_zipper3                    & \\
\hline
\end{tabular}
\end{table}

\noindent The extension defines 42 subclasses in total: 24 single-tag, 16 two-tag, 2 three-tag.

\subsection{Anomalous and normal set construction}
\label{app:perpart:cardinalities}

For each subclass, Table~\ref{tab:perpart_cardinalities} reports the cardinalities $|\mathcal{A}(T_s)|$, $|\mathcal{N}_2(T_s)|$, and the number of excluded images. The normal set under EV1, $\mathcal{N}_1$, contains only the defect-free test images of the class and is therefore constant across subclasses of the same class; per-class $|\mathcal{N}_1|$ values are listed in the per-class header row. Images are excluded from a subclass evaluation when $T_i \cap T_s \neq \emptyset$ and $T_i \not\subseteq T_s$, i.e., when the defect partially overlaps the instructed component while also affecting unrelated components, since such images admit no unambiguous label under either setting.
Counts are computed deterministically from the released annotation file and MVTec~AD's original test split by the reference loader script.

\begin{table}[h]
\centering
\caption{Per-subclass cardinalities under Equation~\eqref{eq:subclass}. $|\mathcal{N}_1|$ (defect-free test images) is class-level and shown in each header row; $|\mathcal{A}|$, $|\mathcal{N}_2|$, and the excluded count are per subclass; the last column gives the total test images of the class.}
\label{tab:perpart_cardinalities}
\footnotesize
\begin{tabular}{llcccc}
\hline
\textbf{Class} & \textbf{$T_s$} & $|T_s|$ & $|\mathcal{A}|$ & $|\mathcal{N}_2|$ & \textbf{Excluded} \\
\hline
\multicolumn{6}{l}{\emph{cable} \quad ($|\mathcal{N}_1| = 58$, total test images $= 92$)} \\
& cable1                       & 1 & 21 & 66 & 5  \\
& cable2                       & 1 & 17 & 58 & 17 \\
& cable3                       & 1 & 17 & 64 & 11 \\
& cable4                       & 1 & 15 & 64 & 13 \\
& cable1\_cable2               & 2 & 2  & 34 & 56 \\
& cable1\_cable3               & 2 & 3  & 41 & 48 \\
& cable2\_cable3               & 2 & 4  & 36 & 52 \\
& cable2\_cable4               & 2 & 9  & 41 & 42 \\
& cable3\_cable4               & 2 & 2  & 40 & 50 \\
& cable2\_cable3\_cable4       & 3 & 2  & 21 & 69 \\
\hline
\multicolumn{6}{l}{\emph{capsule} \quad ($|\mathcal{N}_1| = 23$, total test images $= 109$)} \\
& capsule1                     & 1 & 31 & 66 & 12 \\
& capsule2                     & 1 & 23 & 81 & 5  \\
& capsule3                     & 1 & 38 & 54 & 17 \\
& capsule1\_capsule3           & 2 & 12 & 23 & 74 \\
& capsule2\_capsule3           & 2 & 5  & 31 & 73 \\
\hline
\multicolumn{6}{l}{\emph{hazelnut} \quad ($|\mathcal{N}_1| = 40$, total test images $= 70$)} \\
& hazelnut1                    & 1 & 59 & 1  & 10 \\
& hazelnut2                    & 1 & 1  & 59 & 10 \\
& hazelnut1\_hazelnut2         & 2 & 10 & 0  & 60 \\
\hline
\multicolumn{6}{l}{\emph{metal\_nut} \quad ($|\mathcal{N}_1| = 22$, total test images $= 93$)} \\
& metal\_nut1                  & 1 & 24 & 7  & 62 \\
& metal\_nut2                  & 1 & 7  & 24 & 62 \\
& metal\_nut1\_metal\_nut2     & 2 & 62 & 0  & 31 \\
\hline
\multicolumn{6}{l}{\emph{pill} \quad ($|\mathcal{N}_1| = 26$, total test images $= 141$)} \\
& pill1                        & 1 & 104 & 19  & 18  \\
& pill2                        & 1 & 19  & 104 & 18  \\
& pill1\_pill2                 & 2 & 18  & 0   & 123 \\
\hline
\multicolumn{6}{l}{\emph{screw} \quad ($|\mathcal{N}_1| = 41$, total test images $= 119$)} \\
& screw1                       & 1 & 70 & 49 & 0 \\
& screw2                       & 1 & 24 & 95 & 0 \\
& screw3                       & 1 & 25 & 94 & 0 \\
\hline
\multicolumn{6}{l}{\emph{toothbrush} \quad ($|\mathcal{N}_1| = 12$, total test images $= 30$)} \\
& toothbrush1                                  & 1 & 3  & 23 & 4  \\
& toothbrush2                                  & 1 & 6  & 6  & 18 \\
& toothbrush3                                  & 1 & 3  & 10 & 17 \\
& toothbrush1\_toothbrush2                     & 2 & 1  & 3  & 26 \\
& toothbrush2\_toothbrush3                     & 2 & 14 & 3  & 13 \\
& toothbrush1\_toothbrush2\_toothbrush3        & 3 & 3  & 0  & 27 \\
\hline
\multicolumn{6}{l}{\emph{transistor} \quad ($|\mathcal{N}_1| = 60$, total test images $= 100$)} \\
& transistor1                  & 1 & 10 & 20 & 10 \\
& transistor2                  & 1 & 20 & 10 & 10 \\
& transistor1\_transistor2     & 2 & 10 & 0  & 30 \\
\hline
\multicolumn{6}{l}{\emph{zipper} \quad ($|\mathcal{N}_1| = 32$, total test images $= 120$)} \\
& zipper1                      & 1 & 17 & 95 & 8  \\
& zipper2                      & 1 & 74 & 37 & 9  \\
& zipper3                      & 1 & 17 & 96 & 7  \\
& zipper1\_zipper2             & 2 & 5  & 17 & 98 \\
& zipper1\_zipper3             & 2 & 3  & 74 & 43 \\
& zipper2\_zipper3             & 2 & 4  & 17 & 99 \\
\hline
\end{tabular}
\end{table}

\subsection{Annotation protocol}
\label{app:perpart:protocol}

A single experienced annotator performed all per-image tagging. Each defective test image was assigned the minimal tag set $T_i$ such that every connected component of the ground-truth anomaly mask lies within the union of the regions named by tags in $T_i$. Tags are class-specific and were defined a priori from the visual decomposition of each object (Section~\ref{app:perpart:tags}); the annotator selected tags from this fixed inventory and did not introduce new ones. The ground-truth pixel-level masks from MVTec~AD were used as the spatial reference and were not modified.

\section{Assembled Panel Dataset: Full Specification}
\label{app:assembled_panel}

This appendix specifies the industrial context, acquisition setup, anomaly taxonomy, defect creation protocol, per-category statistics, annotation protocol, and per-anomaly-type prompt protocol of the Assembled Panel dataset.

\subsection{Industrial context}
\label{app:ap:context}

The dataset captures assembled electrical panels for electric-vehicle charging stations, supplied by an industrial partner specialized in the production of such panels. Each panel contains a fixed layout of printed circuit boards (PCBs), power connectors (left and right), EV-charging ports (left and right), current sensors, power and signal cables routed between connectors and PCBs, warning and identification labels, mounting screws with rubber rings, and red protective caps near the EV ports. The anomaly types included in the dataset were selected jointly with domain experts as those most frequent in production and most critical to detect, ensuring that the released data reflects realistic inspection priorities.

\subsection{Acquisition setup}
\label{app:ap:acquisition}

Images were acquired in a single session inside a custom photo box with a reflective silver inner coating that diffuses illumination and a black base panel that minimizes specular returns toward the camera. Panel positioning is enforced by a black perforated board with three locating pins (two long pins at the left corners, one medium pin at the right-middle corner near the second EV port), which constrains in-plane translation and rotation across acquisitions. A protective plastic white base of 14--15\,mm height is interposed between the perforated board and the panel under inspection.

Acquisition was performed with a single Intel RealSense Depth Camera D435i (firmware 5.16.0.1) mounted on a custom-built adjustable stand and operated in RGB mode at $1920 \times 1080$ resolution with PNG encoding. The final configuration (Setup~5 of the acquisition protocol) places the camera in a top-down configuration: stand height 64\,cm measured from the black base to the top of the L-shaped support, camera-to-right-stand distance 52.7\,cm, camera holder fixed by a bottom screw on the outer edge and in direct contact with the metal cross-bar so as to face the panel vertically from above. Illumination is provided exclusively by a manual bottom light at intensity level~3; no top illumination is used in the final setup. Four alternative setups (two-camera oblique views with different stand and lighting configurations) were evaluated during protocol design and discarded in favor of the single top-down view, which provides unobstructed visibility of PCBs, connectors, and cable routing.

To prevent on-camera processing from masking visually subtle anomalies, post-processing and auto-exposure were disabled and all sensor controls were fixed across the session through a standardized configuration file. The salient settings, released with the dataset, are: manual color exposure $1250$\,$\mu$s, color gain $20$, gamma $190$, manual white balance $4620$\,K, saturation $70$, contrast $50$, sharpness $50$, brightness $-1$, hue $-1$, backlight compensation $0$, power-line-frequency filter disabled. The auto-exposure setpoint ($1536$) is reported for completeness but is inactive because auto-exposure is off. The fixed-viewpoint, fixed-illumination, sensor-locked acquisition is the prerequisite for the dataset's central methodological claim: visual variability across acquisitions is attributable to the panel content (defect-free vs.\ anomalous), not to scene factors. The full configuration JSON and the acquisition-protocol document accompany the data release.

\subsection{Anomaly taxonomy}
\label{app:ap:taxonomy}

The taxonomy comprises fourteen anomaly categories defined jointly with domain experts and grouped by reversibility into three types: type~0 (non-reversible damage requiring component replacement), type~1 (reversible defects whose repair is time-consuming), and type~2 (reversible defects with minimal restoration effort). Table~\ref{tab:ap_taxonomy} lists each category with its identifier, descriptive name, reversibility type, and expected visual signature; representative images per category are included in the release. Inspection priority follows the reversibility type itself: a higher type number indicates lower production-restoration cost, and type~0 is the highest-severity, non-recoverable case.

\begin{table}[h]
\centering
\caption{Anomaly taxonomy of the Assembled Panel dataset: fourteen categories with reversibility type ($0$/$1$/$2$) and expected visual signature.}
\label{tab:ap_taxonomy}
\small
\begin{tabular}{cp{0.30\textwidth}cp{0.44\textwidth}}
\hline
\textbf{ID} & \textbf{Category} & \textbf{Type} & \textbf{Visual signature} \\
\hline
A1  & swapped power-cable colors                    & 1 & color permutation at left/right power ports \\
A2  & missing label                                 & 2 & absent identification sticker on PCB black box \\
A3  & missing rubber ring near mounting screw       & 2 & absent ring around panel-mounting screw head \\
A4  & missing power-connection wire(s)              & 1 & absent wire at top/bottom power port \\
A5  & missing screw(s)                              & 2 & absent screw at panel or PCB mounting position \\
A6  & incorrect wire routing through current sensor & 1 & wire count through right current sensor differs from two \\
A7  & scratched or broken PCB                       & 0 & surface scratch, fracture, or missing component on PCB \\
A8  & swapped large cables in EV ports              & 1 & green/blue large cables exchanged between left/right EV ports \\
A9  & swapped small cables in EV ports              & 1 & orange/white small cables exchanged between left/right EV ports \\
A10 & swapped same-color IO power cables            & 1 & upper/lower same-color power cables exchanged at terminal block \\
A11 & misplaced label                               & 2 & identification label present but in non-standard position \\
A12 & missing red cap                               & 2 & absent red protective cap near EV port \\
A13 & missing wire(s) on EV port connections        & 1 & absent wire at left/right EV port connection \\
A14 & unplugged wire(s) on current sensor           & 1 & disconnected wire at the right current sensor \\
\hline
\end{tabular}
\end{table}

\subsection{Defect creation protocol}
\label{app:ap:creation}

A single experienced worker induced each anomalous sample according to category-specific procedures defined with the domain experts, choosing per-category sample counts to reflect both inspection-priority weight and physical-cost constraints. Reversibility determined which categories could yield multiple samples on a single panel: type~1 and type~2 defects are reversible and were induced sequentially on shared panels (e.g., cables unplugged and replugged, screws removed and refitted, labels removed and reapplied), while type~0 defects (scratched or broken PCB) are non-reversible and required dedicated specimens, which constrained the number of acquirable samples for this category.

\subsection{Per-category statistics}
\label{app:ap:statistics}
The dataset contains 99 defect-free images and 53 anomalous images at $1920 \times 1080$, all acquired in the single session described in Section~\ref{app:ap:acquisition}. Of the 53 anomalous images, 14 contain more than one defect category; the per-category sum (69) therefore exceeds the number of anomalous images (53). Table~\ref{tab:ap_stats} reports the per-category counts; Table~\ref{tab:ap_multidefect} enumerates the ten distinct combinations that occur across these 14 multi-defect images.

\begin{table}[h]
\centering
\caption{Per-category anomalous-image counts. \emph{Multi-defect} gives, per category, how many of its images also carry another defect category; totals are reported at the bottom.}
\label{tab:ap_stats}
\small
\begin{tabular}{lcc}
\hline
\textbf{Category} & \textbf{Anomalous} & \textbf{Multi-defect} \\
\hline
A1 swapped power-cable colors                 & 5 & 2 \\
A2 missing label                              & 6 & 4 \\
A3 missing rubber ring                        & 6 & 5 \\
A4 missing/unplugged power-connection wires   & 3 & 2 \\
A5 missing screws                             & 5 & 0 \\
A6 incorrect wire routing through sensor      & 5 & 1 \\
A7 scratched/broken PCB                       & 5 & 1 \\
A8 swapped large cables in EV ports           & 7 & 2 \\
A9 swapped small cables in EV ports           & 5 & 3 \\
A10 swapped same-color IO power cables        & 7 & 3 \\
A11 misplaced label                           & 5 & 0 \\
A12 missing red cap                           & 7 & 4 \\
A13 missing wires on EV port connections      & 1 & 1 \\
A14 unplugged wires on sensor                 & 2 & 1 \\
\hline
\textbf{Sum of per-category counts} & 69 & 29 \\
\textbf{Total anomalous images}     & 53 & --- \\
\textbf{Single-defect images}       & 39 & --- \\
\textbf{Multi-defect images}        & 14 & --- \\
\textbf{Total defect-free images}   & 99 & --- \\
\hline
\end{tabular}
\end{table}

\begin{table}[h]
\centering
\caption{Category combinations occurring in the multi-defect images, with the number of images per combination.}
\label{tab:ap_multidefect}
\small
\begin{tabular}{cll}
\hline
\textbf{\#} & \textbf{Combination} & \textbf{Count} \\
\hline
1  & A3, A10, A12 & 1 \\
2  & A3, A12      & 1 \\
3  & A6, A13      & 1 \\
4  & A10, A12     & 2 \\
5  & A1, A2       & 2 \\
6  & A2, A9       & 2 \\
7  & A3, A4       & 2 \\
8  & A3, A9       & 1 \\
9  & A7, A8       & 1 \\
10 & A8, A14      & 1 \\
\hline
\textbf{Total} & --- & 14 \\
\hline
\end{tabular}
\end{table}

\subsection{Pixel-level annotation protocol}
\label{app:ap:annotation}

Pixel-level annotations were produced in Label Studio at the original $1920 \times 1080$ resolution by a single experienced annotator who performed all per-image work. Defects with well-defined contours (missing components, broken parts, structural anomalies) were outlined as polygons and rasterized to binary masks; diffuse defects such as scratches were annotated directly as masks. For multi-defect images, the released archive provides one mask per defect category present in the image together with a combined mask covering all defects.

\subsection{Per-anomaly-type prompt protocol}
\label{app:ap:prompts}

Scenario~3 evaluates two prompting modes: \textit{Standard~AD}, a single prompt across all categories at four levels of descriptive richness, and \textit{Per-Error~AD}, a category-specific prompt naming the relevant component and the expected failure mode. The full prompt sets for all three models, in both modes, are listed by model in Appendix~\ref{app:prompts_scen3}.

\subsection{Extended Assembled Panel results}
\label{app:ap:extended}

Tables~\ref{tab:panel_224_direct_resize} and~\ref{tab:panel_224_pad_to_square} report the full Standard~AD sweep on Assembled Panel at $224\times224$, across all prompt variants and supervision regimes, for the two input-handling strategies (Direct Resize and Pad to Square). The best configuration per model, summarized in Table~\ref{tab:model_config_results}, is drawn from these.

\begin{table}[h]
\centering
\caption{Assembled Panel, Standard~AD at $224 \times 224$ with Direct Resize. I-AUROC, P-AUROC, and AUPRO per model and prompt variant.}
\label{tab:panel_224_direct_resize}
\small
\begin{tabular}{llccc}
\hline
\textbf{Model} & \textbf{Prompt variant}
& \textbf{I-AUROC} & \textbf{P-AUROC} & \textbf{AUPRO} \\
\hline
AnomalyGPT
& Very Detailed & 71.7 & 82.6 & 22.6 \\
(0-shot) & Detailed      & 71.7 & 82.6 & 22.6 \\
& Simple        & 78.6 & 81.6 & 22.8 \\
& Minimal       & 78.6 & 81.6 & 22.8 \\
\hline
AnomalyGPT
& Very Detailed        & 65.6 & 76.1 & 17.8 \\
(1-shot) & Minimal       & 65.6 & 76.1 & 17.8 \\
\hline
LogSAD
& Minimal       & 40.9 & 81.2 & 35.5 \\
(1-shot) & Simple        & 40.9 & 81.2 & 35.5 \\
& Very Detailed & 40.9 & 81.2 & 35.5 \\
\hline
LogSAD
& Very Detailed & 45.1 & 78.7 & 35.5 \\
(full-shot) & Simple        & 45.1 & 78.7 & 35.5 \\
& Minimal       & 45.1 & 78.7 & 35.5 \\
\hline
AA-CLIP
& Very Detailed & 21.2 & 87.5 & 65.3 \\
(0-shot) & Simple        & 20.5 & 87.2 & 64.8 \\
& Minimal       & 23.7 & 87.9 & 66.2 \\
\hline
\end{tabular}
\end{table}

\begin{table}[h]
\centering
\caption{Assembled Panel, Standard~AD at $224 \times 224$ with Pad to Square. I-AUROC, P-AUROC, and AUPRO per model and prompt variant.}
\label{tab:panel_224_pad_to_square}
\small
\begin{tabular}{llccc}
\hline
\textbf{Model} & \textbf{Prompt variant}
& \textbf{I-AUROC} & \textbf{P-AUROC} & \textbf{AUPRO} \\
\hline
AnomalyGPT
& Very Detailed & 61.3 & 77.0 & 20.4 \\
(0-shot) & Minimal       & 61.0 & 81.5 & 22.9 \\
\hline
AnomalyGPT
& Very Detailed        & 58.4 & 67.1 & 10.2 \\
(1-shot) & Minimal       & 58.4 & 67.1 & 10.2 \\
\hline
LogSAD
& Very Detailed & 49.0 & 71.3 & 35.6 \\
(1-shot) & Simple        & 49.0 & 71.3 & 35.6 \\
& Minimal       & 49.0 & 71.3 & 35.6 \\
\hline
LogSAD
& Very Detailed & 62.6 & 73.1 & 44.5 \\
(full-shot) & Simple        & 62.6 & 73.1 & 44.5 \\
& Minimal       & 62.6 & 73.1 & 44.5 \\
\hline
AA-CLIP
& Very Detailed & 31.5 & 84.4 & 55.9 \\
(0-shot) & Simple        & 31.8 & 84.1 & 55.2 \\
& Minimal       & 33.2 & 84.9 & 56.6 \\
\hline
\end{tabular}
\end{table}


\section{Scenario 1 Prompts}
\label{app:prompts_scen1}
This appendix reports verbatim every textual input used in the Scenario~1 evaluation (Section~\ref{sec:res_scen1}) on MVTec~AD, organized by model. AnomalyGPT consumes a free-form prompt, listed per class for each of the five variants; LogSAD and AA-CLIP consume keywords inserted into a fixed template ensemble, listed per class for each keyword form. The prompts are reported as used, without templating.

\subsection{AnomalyGPT}
\label{app:prompts_scen1:anomalygpt}
Five prompt variants are evaluated per class. \textit{Original Description} states the object and a flaw enumeration; \textit{New Description} is a free paraphrase; \textit{No Anomaly Description} keeps the object but drops the flaw enumeration; \textit{No Normality Description} keeps the flaw enumeration but drops the normality assertion; \textit{Minimal} keeps only the question form.

\begin{promptbox}
\textbf{Minimal (all classes).}\\[2pt]
``Is there any anomaly in the image?''
\end{promptbox}

\begin{promptbox}
\textbf{Original Description.}\\[2pt]
\textbf{Bottle.}~``This is a photo of a bottle for anomaly detection, which should be round, without any damage, flaw, defect, scratch, hole or broken part. Is there any anomaly in the image?'' \\[3pt]
\textbf{Cable.}~``This is a photo of three cables for anomaly detection, cables cannot be missed or swapped, which should be without any damage, flaw, defect, scratch, hole or broken part. Is there any anomaly in the image?'' \\[3pt]
\textbf{Capsule.}~``This is a photo of a capsule for anomaly detection, which should be black and orange, with print '500', without any damage, flaw, defect, scratch, hole or broken part. Is there any anomaly in the image?'' \\[3pt]
\textbf{Carpet.}~``This is a photo of carpet for anomaly detection, which should be without any damage, flaw, defect, scratch, hole or broken part. Is there any anomaly in the image?'' \\[3pt]
\textbf{Grid.}~``This is a photo of grid for anomaly detection, which should be without any damage, flaw, defect, scratch, hole or broken part. Is there any anomaly in the image?'' \\[3pt]
\textbf{Hazelnut.}~``This is a photo of a hazelnut for anomaly detection, which should be without any damage, flaw, defect, scratch, hole or broken part. Is there any anomaly in the image?'' \\[3pt]
\textbf{Leather.}~``This is a photo of leather for anomaly detection, which should be brown and without any damage, flaw, defect, scratch, hole or broken part. Is there any anomaly in the image?'' \\[3pt]
\textbf{Metal Nut.}~``This is a photo of a metal nut for anomaly detection, which should be without any damage, flaw, defect, scratch, hole or broken part, and shouldn't be flipped. Is there any anomaly in the image?'' \\[3pt]
\textbf{Pill.}~``This is a photo of a pill for anomaly detection, which should be white, with print 'FF' and red patterns, without any damage, flaw, defect, scratch, hole or broken part. Is there any anomaly in the image?'' \\[3pt]
\textbf{Screw.}~``This is a photo of a screw for anomaly detection, which tail should be sharp, and without any damage, flaw, defect, scratch, hole or broken part. Is there any anomaly in the image?'' \\[3pt]
\textbf{Tile.}~``This is a photo of tile for anomaly detection, which should be without any damage, flaw, defect, scratch, hole or broken part. Is there any anomaly in the image?'' \\[3pt]
\textbf{Toothbrush.}~``This is a photo of a toothbrush for anomaly detection, which should be without any damage, flaw, defect, scratch, hole or broken part. Is there any anomaly in the image?'' \\[3pt]
\textbf{Transistor.}~``This is a photo of a transistor for anomaly detection, which should be without any damage, flaw, defect, scratch, hole or broken part. Is there any anomaly in the image?'' \\[3pt]
\textbf{Wood.}~``This is a photo of wood for anomaly detection, which should be brown with patterns, without any damage, flaw, defect, scratch, hole or broken part. Is there any anomaly in the image?'' \\[3pt]
\textbf{Zipper.}~``This is a photo of a zipper for anomaly detection, which should be without any damage, flaw, defect, scratch, hole or broken part. Is there any anomaly in the image?''
\end{promptbox}

\begin{promptbox}
\textbf{New Description (paraphrase).}\\[2pt]
\textbf{Bottle.}~``A close-up view of a single bottle designed for quality control. The object should exhibit a smooth, uninterrupted round form with no visual indicators of wear, damage, or manufacturing errors. Is there any anomaly in the image?'' \\[3pt]
\textbf{Cable.}~``Three distinct cables appear in this scene, each expected to lie in proper order. The inspection focuses on ensuring that none are missing, misplaced, or show visible surface damage like cuts or abrasions. Is there any anomaly in the image?'' \\[3pt]
\textbf{Capsule.}~``Captured here is a pharmaceutical capsule featuring black and orange segments with the marking '500'. Any inconsistency in color, missing print, or physical deformity would indicate a defect. Is there any anomaly in the image?'' \\[3pt]
\textbf{Carpet.}~``An area of carpet is shown, where attention should be given to texture uniformity and surface integrity. Defects like punctures, fiber disruptions, or irregular patterns are not acceptable. Is there any anomaly in the image?'' \\[3pt]
\textbf{Grid.}~``This image highlights a metallic grid structure. Gaps, broken lines, or any visible structural inconsistencies would classify as anomalies during inspection. Is there any anomaly in the image?'' \\[3pt]
\textbf{Hazelnut.}~``A single hazelnut rests at the center of this frame. Inspectors should check for a clean, unbroken shell with no cracks, holes, or discoloration that might signal a flaw. Is there any anomaly in the image?'' \\[3pt]
\textbf{Leather.}~``A stretched piece of brown leather is displayed. Consistent surface texture and color are required; scratches, tears, or stains would signal an anomaly. Is there any anomaly in the image?'' \\[3pt]
\textbf{Metal Nut.}~``The photograph shows a metal fastening nut. It should be oriented correctly, with its entire surface free from dents, scratches, missing edges, or other deformities. Is there any anomaly in the image?'' \\[3pt]
\textbf{Pill.}~``This frame contains a single pill marked with 'FF' and decorated with red designs on a white background. Any smudging, shape distortion, or surface damage is considered a defect. Is there any anomaly in the image?'' \\[3pt]
\textbf{Screw.}~``The focus here is on a single metal screw. Its pointed end must remain sharp, and no part of its surface should exhibit bends, scratches, or manufacturing errors. Is there any anomaly in the image?'' \\[3pt]
\textbf{Tile.}~``A square tile is positioned at the center of this image. Inspectors are to look for surface consistency, with no chips, holes, or fractures present. Is there any anomaly in the image?'' \\[3pt]
\textbf{Toothbrush.}~``Presented here is a toothbrush. Any broken bristles, cracks in the handle, missing parts, or visible marks would categorize it as defective. Is there any anomaly in the image?'' \\[3pt]
\textbf{Transistor.}~``A small electronic transistor sits against a plain background. The pins must remain straight and intact, with no evidence of bending, scratches, or material loss. Is there any anomaly in the image?'' \\[3pt]
\textbf{Wood.}~``This is a wooden surface sample. Natural grain patterns should appear uninterrupted, and no cuts, holes, color patches, or texture defects should exist. Is there any anomaly in the image?'' \\[3pt]
\textbf{Zipper.}~``The subject here is a clothing zipper. Teeth alignment must be perfect, and no separations, broken segments, or material damage should appear along its length. Is there any anomaly in the image?''
\end{promptbox}

\begin{promptbox}
\textbf{No Anomaly Description (flaw enumeration removed).}\\[2pt]
\textbf{Bottle.}~``This is a photo of a bottle for anomaly detection, which should be round. Is there any anomaly in the image?'' \\[3pt]
\textbf{Cable.}~``This is a photo of three cables for anomaly detection. Is there any anomaly in the image?'' \\[3pt]
\textbf{Capsule.}~``This is a photo of a capsule for anomaly detection, which should be black and orange, with print '500'. Is there any anomaly in the image?'' \\[3pt]
\textbf{Carpet.}~``This is a photo of carpet for anomaly detection. Is there any anomaly in the image?'' \\[3pt]
\textbf{Grid.}~``This is a photo of grid for anomaly detection. Is there any anomaly in the image?'' \\[3pt]
\textbf{Hazelnut.}~``This is a photo of a hazelnut for anomaly detection. Is there any anomaly in the image?'' \\[3pt]
\textbf{Leather.}~``This is a photo of leather for anomaly detection, which should be brown. Is there any anomaly in the image?'' \\[3pt]
\textbf{Metal Nut.}~``This is a photo of a metal nut for anomaly detection. Is there any anomaly in the image?'' \\[3pt]
\textbf{Pill.}~``This is a photo of a pill for anomaly detection, which should be white, with print 'FF' and red patterns. Is there any anomaly in the image?'' \\[3pt]
\textbf{Screw.}~``This is a photo of a screw for anomaly detection, which tail should be sharp. Is there any anomaly in the image?'' \\[3pt]
\textbf{Tile.}~``This is a photo of tile for anomaly detection. Is there any anomaly in the image?'' \\[3pt]
\textbf{Toothbrush.}~``This is a photo of a toothbrush for anomaly detection. Is there any anomaly in the image?'' \\[3pt]
\textbf{Transistor.}~``This is a photo of a transistor for anomaly detection. Is there any anomaly in the image?'' \\[3pt]
\textbf{Wood.}~``This is a photo of wood for anomaly detection, which should be brown with patterns. Is there any anomaly in the image?'' \\[3pt]
\textbf{Zipper.}~``This is a photo of a zipper for anomaly detection. Is there any anomaly in the image?''
\end{promptbox}

\begin{promptbox}
\textbf{No Normality Description (normality assertion removed).}\\[2pt]
\textbf{Bottle.}~``This is a photo of a bottle for anomaly detection, without any damage, flaw, defect, scratch, hole or broken part. Is there any anomaly in the image?'' \\[3pt]
\textbf{Cable.}~``This is a photo of three cables for anomaly detection, cables cannot be missed or swapped, which should be without any damage, flaw, defect, scratch, hole or broken part. Is there any anomaly in the image?'' \\[3pt]
\textbf{Capsule.}~``This is a photo of a capsule for anomaly detection, which should be without any damage, flaw, defect, scratch, hole or broken part. Is there any anomaly in the image?'' \\[3pt]
\textbf{Carpet.}~``This is a photo of carpet for anomaly detection, which should be without any damage, flaw, defect, scratch, hole or broken part. Is there any anomaly in the image?'' \\[3pt]
\textbf{Grid.}~``This is a photo of grid for anomaly detection, which should be without any damage, flaw, defect, scratch, hole or broken part. Is there any anomaly in the image?'' \\[3pt]
\textbf{Hazelnut.}~``This is a photo of a hazelnut for anomaly detection, which should be without any damage, flaw, defect, scratch, hole or broken part. Is there any anomaly in the image?'' \\[3pt]
\textbf{Leather.}~``This is a photo of leather for anomaly detection, which should be without any damage, flaw, defect, scratch, hole or broken part. Is there any anomaly in the image?'' \\[3pt]
\textbf{Metal Nut.}~``This is a photo of a metal nut for anomaly detection, which should be without any damage, flaw, defect, scratch, hole or broken part, and shouldn't be flipped. Is there any anomaly in the image?'' \\[3pt]
\textbf{Pill.}~``This is a photo of a pill for anomaly detection, which should be without any damage, flaw, defect, scratch, hole or broken part. Is there any anomaly in the image?'' \\[3pt]
\textbf{Screw.}~``This is a photo of a screw for anomaly detection, which should be without any damage, flaw, defect, scratch, hole or broken part. Is there any anomaly in the image?'' \\[3pt]
\textbf{Tile.}~``This is a photo of tile for anomaly detection, which should be without any damage, flaw, defect, scratch, hole or broken part. Is there any anomaly in the image?'' \\[3pt]
\textbf{Toothbrush.}~``This is a photo of a toothbrush for anomaly detection, which should be without any damage, flaw, defect, scratch, hole or broken part. Is there any anomaly in the image?'' \\[3pt]
\textbf{Transistor.}~``This is a photo of a transistor for anomaly detection, which should be without any damage, flaw, defect, scratch, hole or broken part. Is there any anomaly in the image?'' \\[3pt]
\textbf{Wood.}~``This is a photo of wood for anomaly detection, which should be without any damage, flaw, defect, scratch, hole or broken part. Is there any anomaly in the image?'' \\[3pt]
\textbf{Zipper.}~``This is a photo of a zipper for anomaly detection, which should be without any damage, flaw, defect, scratch, hole or broken part. Is there any anomaly in the image?''
\end{promptbox}

\subsection{LogSAD}
\label{app:prompts_scen1:logsad}
LogSAD's textual input acts only at the foreground stage: a foreground keyword and the background keyword \textit{background} are each inserted into a fixed ensemble of natural-language templates, encoded with CLIP, and averaged into a single text embedding used to separate foreground from background. We test three foreground keyword forms per class, listed in Table~\ref{tab:keywords_logsad}.

\begin{table}[h]
\caption{LogSAD foreground keywords on MVTec~AD; the background keyword is \textit{background} in all cases.}
\centering
\small
\begin{tabular}{llll}
\toprule
Class & Simple & Very Detailed & Minimal \\
\midrule
Bottle & bottle & glass bottle & object \\
Cable & cable & electric cable & object \\
Capsule & capsule & medicine capsule & object \\
Carpet & carpet & floor mat & texture \\
Grid & grid & metal grid & texture \\
Hazelnut & hazelnut & nut & object \\
Leather & leather & leather material & texture \\
Metal Nut & metal nut & hex nut & object \\
Pill & pill & tablet & object \\
Screw & screw & metal screw & object \\
Tile & tile & ceramic tile & texture \\
Toothbrush & toothbrush & brush & object \\
Transistor & transistor & electronic component & object \\
Wood & wood & wooden surface & texture \\
Zipper & zipper & clothing zipper & object \\
\bottomrule
\end{tabular}
\label{tab:keywords_logsad}
\end{table}

\subsection{AA-CLIP}
\label{app:prompts_scen1:aaclip}
AA-CLIP inserts a class keyword into normal and abnormal state patterns (for example ``a \{\}'', ``a clean \{\}'' for the normal state and ``a damaged \{\}'', ``a \{\} with defect'' for the abnormal state), which are then embedded into a fixed template ensemble (for example ``\{\}.'', ``a photo of \{\}.'', ``this is \{\}.''); the normalized embeddings are averaged into one normal and one abnormal text anchor. We test three keyword forms per class, listed in Table~\ref{tab:keywords_aaclip}.

\begin{table}[h]
\caption{AA-CLIP keywords on MVTec~AD across the three evaluated forms.}
\centering
\small
\begin{tabular}{l p{5.2cm} l l}
\toprule
Class & Very Detailed & Simple & Minimal \\
\midrule
Bottle & dark bottle & bottle & object \\
Cable & top view of three cables & cable & object \\
Capsule & black and orange capsule & capsule & object \\
Carpet & gray carpet & carpet & object \\
Grid & metal or plastic mesh & grid & object \\
Hazelnut & single brown hazelnut & hazelnut & object \\
Leather & brown leather & leather & object \\
Metal Nut & metal nut which has four notched edges & metal\_nut & object \\
Pill & oval white pill with small red speckles and the letters 'FF' engraved & pill & object \\
Screw & screw & screw & object \\
Tile & speckled tile surface & tile & object \\
Toothbrush & toothbrush head & toothbrush & object \\
Transistor & a three-legged transistor placed vertically & transistor & object \\
Wood & wood surface & wood & object \\
Zipper & a black zipper & zipper & object \\
\bottomrule
\end{tabular}
\label{tab:keywords_aaclip}
\end{table}

\section{Scenario 2 Prompts}
\label{app:prompts_scen2}
This appendix reports the per-subclass textual inputs used in the Scenario~2 (per-part) evaluation on the MVTec~AD extension (Section~\ref{sec:res_scen2}), organized by model. AnomalyGPT consumes a free-form instruction; LogSAD splits foreground from background using a foreground (F) and a background (B) keyword set; AA-CLIP inserts a component keyword into its state-conditioned templates. Subclasses follow the tag notation of Appendix~\ref{app:perpart} (single tags, and multi-tag sets joined by an underscore).

\subsection{AnomalyGPT}
\label{app:prompts_scen2:anomalygpt}
For each subclass, AnomalyGPT receives an instruction to inspect only the tagged component(s) and to ignore the rest of the object. To reduce sensitivity to phrasing, three semantically equivalent formulations are written per subclass, and one is drawn uniformly at random per image with a fixed seed. The full per-subclass pools for all nine annotated classes are listed below.

\begin{promptbox}
\textbf{cable1 (light-blue insulation)}\\[2pt]
1. Focus only on examining the light blue insulator of the conductor for any flaws.\\
2. Assess solely the light blue insulation within the conductor for signs of damage.\\
3. Pay close attention only to the insulation layer in the conductor, which is light blue, for any potential issues.
\end{promptbox}

\begin{promptbox}
\textbf{cable2 (green cable)}\\[2pt]
1. Focus solely on the green wire to check for any issues.\\
2. Verify the integrity of the green cable only before proceeding.\\
3. Examine the green cable only in detail to ensure it's in good condition.
\end{promptbox}

\begin{promptbox}
\textbf{cable3 (blue cable)}\\[2pt]
1. Focus solely on the blue cable for any imperfections during inspection.\\
2. Attention must be directed exclusively towards the blue cable when examining for flaws.\\
3. During the examination process, ensure that you closely examine the blue cable only without looking at the other cables.
\end{promptbox}

\begin{promptbox}
\textbf{cable4 (gray cable)}\\[2pt]
1. Focus exclusively on the gray wire when examining it for potential flaws.\\
2. It is crucial to scrutinize solely the gray wire within this setting.\\
3. For the purposes of analysis, concentrate solely on the gray cable in this particular scene.
\end{promptbox}

\begin{promptbox}
\textbf{cable1\_cable2}\\[2pt]
1. For a thorough examination, focus solely on the light blue coating and the green wire in this image.\\
2. When conducting a meticulous inspection, pay close attention to the light-blue insulation and the green cable depicted here.\\
3. To ensure the quality of these components, closely examine only the green wire and the light-blue coating shown in the image.
\end{promptbox}

\begin{promptbox}
\textbf{cable1\_cable3}\\[2pt]
1. Focus solely on the light blue cover and blue wire, looking out for any imperfections or faults.\\
2. Examine closely solely the light-blue insulation layer and the blue cable for any potential issues.\\
3. Pay close attention only to the light blue covering and the blue conductor within the container, searching for signs of defect or malfunction.
\end{promptbox}

\begin{promptbox}
\textbf{cable2\_cable3}\\[2pt]
1. Pay attention solely to the green wire and blue wire in this image for any potential issues.\\
2. Examine carefully only the green wire and the blue wire within the image for any sign of defects.\\
3. Look closely only at the green cable and the blue cable visible in the photo for any flaws or imperfections.
\end{promptbox}

\begin{promptbox}
\textbf{cable2\_cable4}\\[2pt]
1. Examine just the green wire and silver wire for potential flaws.\\
2. Look closely only at the green conductor and gray tube to ensure their quality.\\
3. Check solely the green line and gray pipe for any imperfections.
\end{promptbox}

\begin{promptbox}
\textbf{cable3\_cable4}\\[2pt]
1. Focus solely on examining the blue and gray cables for possible flaws.\\
2. Attention should be confined to inspecting the blue and gray cable lines only for any defects.\\
3. Check only the blue and gray cables for potential issues during the inspection process.
\end{promptbox}

\begin{promptbox}
\textbf{cable2\_cable3\_cable4}\\[2pt]
1. Pay attention solely to the green cable, blue cable, and gray cable when examining the wires.\\
2. Focus on evaluating only the condition of the green cable, blue cable, and gray cable for any faults.\\
3. Examine exclusively the green cable, blue cable, and gray cable for possible defects.
\end{promptbox}

\begin{promptbox}
\textbf{capsule1 (white printed ``500'')}\\[2pt]
1. Ensure to scrutinize solely the 500-word content for any imperfections.\\
2. Check only for flaws in the lines of white text.\\
3. Review specifically the 500-word content marked by white characters.
\end{promptbox}

\begin{promptbox}
\textbf{capsule2 (left black part)}\\[2pt]
1. Take note to scrutinize just the black portion on the left for any signs of imperfection.\\
2. It is crucial to examine solely the black section on the left when assessing for flaws.\\
3. Be vigilant and carefully check only the black area located on the left side for any defects.
\end{promptbox}

\begin{promptbox}
\textbf{capsule3 (right reddish-orange part)}\\[2pt]
1. Examine solely the red/orange section on the right for any flaws.\\
2. Assess just the proper hue of red/orange on the right for any imperfections.\\
3. Check exclusively the accurate right reddish-orange area for any defects.
\end{promptbox}

\begin{promptbox}
\textbf{capsule1\_capsule3}\\[2pt]
1. Examine just the white text and orange side of the pill for any irregularities or imperfections.\\
2. Look closely only at the small white label and the larger reddish-orange portion of the pill for potential issues.\\
3. Focus exclusively on the 500 white writing and the right orange side of the capsule for defects that may be present.
\end{promptbox}

\begin{promptbox}
\textbf{capsule2\_capsule3}\\[2pt]
1. Look closely solely at the left black section and the right reddish-orange portion of the pill for any imperfections.\\
2. Examine the pill by focusing just on the dark left side and the vibrant red right side for any flaws.\\
3. Investigate both ends of the pill, the left black part and the right orange part for any defects or issues.
\end{promptbox}

\begin{promptbox}
\textbf{hazelnut1 (shell)}\\[2pt]
1. Check just the outer casing for any flaws before consumption.\\
2. Examine solely the hard outer shell for imperfections when assessing quality.\\
3. Take a close look only at the exterior shell to ensure it's in good condition.
\end{promptbox}

\begin{promptbox}
\textbf{hazelnut2 (hilum)}\\[2pt]
1. Inspect only the hilum area, do not consider the outer shell.\\
2. Focus solely your attention on the circular scarred base of the nut, ignoring the shell surface.\\
3. Evaluate only the attachment point of the hazelnut (hilum), not the surrounding shell.
\end{promptbox}

\begin{promptbox}
\textbf{hazelnut1\_hazelnut2}\\[2pt]
1. Pay close attention only to the outer casing and the attachment point of the hazelnut, checking for any imperfections.\\
2. Examine solely the shell and hilum of the hazelnut, looking out for any signs of damage or flaw.\\
3. Focus solely on the shell and hilum of the walnut when assessing its quality.
\end{promptbox}

\begin{promptbox}
\textbf{metal\_nut1 (outer hexagonal body)}\\[2pt]
1. Examine only the outer, hexagonal form for any imperfections before use.\\
2. Before incorporating the object into a system, closely observe only its exterior hexagonal shape for potential flaws.\\
3. Prior to employing it, carefully scrutinize exclusively the external hexagonal body for any defects that may hinder functionality.
\end{promptbox}

\begin{promptbox}
\textbf{metal\_nut2 (central threaded hole)}\\[2pt]
1. Focus solely on the central, threaded opening to identify any imperfections.\\
2. Examine the core of the hole only for signs of damage or irregularities.\\
3. Study just the threaded center of the hole for defects before proceeding with further actions.
\end{promptbox}

\begin{promptbox}
\textbf{metal\_nut1\_metal\_nut2}\\[2pt]
1. Examine the exterior hexagonal shape and the core threaded aperture for any imperfections.\\
2. Verify the integrity of the outer hexagonal body as well as the central hole with threading, inspecting them closely.\\
3. Check both the external hexagonal structure and the inner threaded opening for signs of damage or flaw before proceeding.
\end{promptbox}

\begin{promptbox}
\textbf{pill1 (plain surface)}\\[2pt]
1. Carefully examine only the flat side of the pill capsule for any imperfections.\\
2. It's important to thoroughly inspect just the upper surface of the pill cap for any flaws before use.\\
3. Before ingesting, please check only the top part of the pill container for any defects or issues that may affect its performance.
\end{promptbox}

\begin{promptbox}
\textbf{pill2 (embossed letters ``FF'')}\\[2pt]
1. Take a closer look solely at the imprinted `FF' on the pill to ensure its quality.\\
2. Verify only if the embossed letters FF on the pill are flawless for authenticity purposes.\\
3. Check for any imperfections exclusively in the embossed text FF on the pill before consumption.
\end{promptbox}

\begin{promptbox}
\textbf{pill1\_pill2}\\[2pt]
1. Scrutinize solely the flat expanse and engraved characters `FF' on the tablet to look for any flaws.\\
2. Focus your attention purely on the tablet's smooth exterior and imprinted text `FF' when checking for imperfections.\\
3. Pay close attention to the even surface and inscribed letters `FF' on the pill while examining for any defects.
\end{promptbox}

\begin{promptbox}
\textbf{screw1 (threaded shaft)}\\[2pt]
1. Look only at the spiral-threaded section of the screw. Ignore the flat head and the smooth neck.\\
2. Focus solely on the midsection with threads. Do not evaluate the screw head or neck.\\
3. Your attention should be restricted to the part where the screw has helical grooves. Disregard the rest.
\end{promptbox}

\begin{promptbox}
\textbf{screw2 (screw head)}\\[2pt]
1. Examine just the top of the screw for any imperfections.\\
2. Assess exclusively the head of the screw for potential issues.\\
3. Check solely the portion where you would insert a tool into the screw for any flaws.
\end{promptbox}

\begin{promptbox}
\textbf{screw3 (neck)}\\[2pt]
1. Assess only the neck of the screw for any imperfections before proceeding with installation.\\
2. Examine exclusively the shaft without the threads of the screw before using to ensure it's in good condition.\\
3. Check only the screw neck thoroughly for any issues prior to use.
\end{promptbox}

\begin{promptbox}
\textbf{toothbrush1 (plastic base)}\\[2pt]
1. Assess only the quality of the base by checking for any flaws.\\
2. Verify exclusively if the plastic base is in good condition before use.\\
3. Ensure just that the plastic base component is free from damage before proceeding with its use.
\end{promptbox}

\begin{promptbox}
\textbf{toothbrush2 (white bristles)}\\[2pt]
1. Focus solely on the white bristles when evaluating for flaws.\\
2. Examine only the bristles carefully for any imperfections, paying particular attention to the white ones.\\
3. When assessing the toothbrush, concentrate exclusively on the bristles' color and quality specifically focusing on the white bristles.
\end{promptbox}

\begin{promptbox}
\textbf{toothbrush3 (colored bristles)}\\[2pt]
1. Examine solely the colored bristles of a toothbrush for any flaws.\\
2. Assess only the colored bristles on a toothbrush for any imperfections.\\
3. Scrutinize exclusively the toothbrush's colored bristles for any defects during inspection.
\end{promptbox}

\begin{promptbox}
\textbf{toothbrush1\_toothbrush2}\\[2pt]
1. Examine only the white bristles and the base of the brush, focusing on any imperfections.\\
2. Pay close attention just to both the plastic handle and the white bristles for any signs of damage or defects.\\
3. Carefully check solely the white bristles and the plastic base for any flaws before using it.
\end{promptbox}

\begin{promptbox}
\textbf{toothbrush2\_toothbrush3}\\[2pt]
1. Examine solely the white bristles and colorful bristles on the brush for any flaws.\\
2. Look over the white bristles and colorful bristles of the brush only to identify any issues.\\
3. Inspect solely the brush's white bristles and colored bristles to ensure they are in good condition.
\end{promptbox}

\begin{promptbox}
\textbf{toothbrush1\_toothbrush2\_toothbrush3}\\[2pt]
1. Focus on the plastic handle, white bristles, and colored bristles for any imperfections during inspection.\\
2. Evaluate solely the plastic stand, white brushes, and colored brushes to identify any flaws.\\
3. Assess merely the plastic grip, white brush tips, and blue brush heads for any defects before use.
\end{promptbox}

\begin{promptbox}
\textbf{transistor1 (plastic body)}\\[2pt]
1. Focus solely on the plastic casing when checking for imperfections.\\
2. Take a close look only at the plastic exterior when evaluating its condition.\\
3. Look just at the plastic part and nothing else when examining for flaws.
\end{promptbox}

\begin{promptbox}
\textbf{transistor2 (metallic legs)}\\[2pt]
1. Examine solely the metallic legs carefully to identify any imperfections.\\
2. Verify if there are any flaws in the metalic components during inspection.\\
3. Look for any structural issues exclusively on the metallic elements of the object.
\end{promptbox}

\begin{promptbox}
\textbf{transistor1\_transistor2}\\[2pt]
1. Check only the plastic casing and metallic connectors for any flaws.\\
2. Examine just the plastic part and the metal contacts for any imperfections.\\
3. Assess only the plastic cover and the metal leads for any signs of damage.
\end{promptbox}

\begin{promptbox}
\textbf{zipper1 (interior fabric)}\\[2pt]
1. Focus only on the inner texture of the material to identify any imperfections.\\
2. When examining, pay close attention solely to the internal fabric component.\\
3. Check for defects within the inner fabric's structure only.
\end{promptbox}

\begin{promptbox}
\textbf{zipper2 (zipper teeth)}\\[2pt]
1. Focus solely on examining the zipper's teeth for any imperfections.\\
2. Examine only the teeth of the zipper to identify any flaws.\\
3. Assess the condition of the zipper's teeth only, looking for signs of damage.
\end{promptbox}

\begin{promptbox}
\textbf{zipper3 (fabric at the edge)}\\[2pt]
1. Focus solely on examining the fabric at the edge of the item for any flaws.\\
2. Assess just the material specifically located at the rim of the object for any imperfections.\\
3. Examine exclusively the fabric present at the periphery of the item to determine if there are any defects.
\end{promptbox}

\begin{promptbox}
\textbf{zipper1\_zipper2}\\[2pt]
1. Examine closely just the inner fabric and zipper mechanism to identify any flaws.\\
2. Look carefully only at the material inside and the zipper's teeth for any imperfections.\\
3. Assess solely the internal fabric and the zipper's alignment for potential defects.
\end{promptbox}

\begin{promptbox}
\textbf{zipper1\_zipper3}\\[2pt]
1. Examine only the inner material and fabric along the perimeter for any imperfections.\\
2. Evaluate just the textile's internal and border areas for any flaws.\\
3. Scrutinize the inside fabric and edge fabric only for potential issues.
\end{promptbox}

\begin{promptbox}
\textbf{zipper2\_zipper3}\\[2pt]
1. Assess just the zipper fastenings and the border fabric for any flaws.\\
2. Carefully examine only the teeth of the zipper and the border material at the edge for any damage or issues.\\
3. Focus solely on inspecting the zipper's engagement points and the fabric at the edge for any defects.
\end{promptbox}

\subsection{LogSAD}
\label{app:prompts_scen2:logsad}
LogSAD performs per-part inspection through the foreground-masking extension: a foreground keyword set (F) names the instructed component(s) and a background keyword set (B) names the remaining components plus the generic term \textit{background}. Each keyword is expanded into the template ensemble, encoded with CLIP, and averaged; the resulting embeddings drive the foreground mask of Equations~\ref{eq:fg_mask}--\ref{eq:masked_map}. The per-subclass F and B sets are listed below, one box per class.

\begin{promptbox}
\textbf{Cable.}\\[2pt]
\textbf{cable1.} F: light blue outer insulation. B: green inner cable, blue inner cable, gray inner cable, background.\\[3pt]
\textbf{cable2.} F: green inner cable. B: light blue outer insulation, blue inner cable, gray inner cable, background.\\[3pt]
\textbf{cable3.} F: blue inner cable. B: light blue outer insulation, green inner cable, gray inner cable, background.\\[3pt]
\textbf{cable4.} F: gray inner cable. B: light blue outer insulation, green inner cable, blue inner cable, background.\\[3pt]
\textbf{cable1\_cable2.} F: light blue outer insulation, green inner cable. B: blue inner cable, gray inner cable, background.\\[3pt]
\textbf{cable1\_cable3.} F: light blue outer insulation, blue inner cable. B: green inner cable, gray inner cable, background.\\[3pt]
\textbf{cable2\_cable3.} F: green inner cable, blue inner cable. B: light blue outer insulation, gray inner cable, background.\\[3pt]
\textbf{cable2\_cable4.} F: green inner cable, gray inner cable. B: light blue outer insulation, blue inner cable, background.\\[3pt]
\textbf{cable3\_cable4.} F: blue inner cable, gray inner cable. B: light blue outer insulation, green inner cable, background.\\[3pt]
\textbf{cable2\_cable3\_cable4.} F: green inner cable, blue inner cable, gray inner cable. B: light blue outer insulation, background.
\end{promptbox}

\begin{promptbox}
\textbf{Capsule.}\\[2pt]
\textbf{capsule1.} F: white printed number 500. B: black gelatin capsule, orange gelatin capsule, background.\\[3pt]
\textbf{capsule2.} F: black gelatin capsule. B: white printed number 500, orange gelatin capsule, background.\\[3pt]
\textbf{capsule3.} F: orange gelatin capsule. B: white printed number 500, black gelatin capsule, background.\\[3pt]
\textbf{capsule1\_capsule3.} F: white printed number 500, orange gelatin capsule. B: black gelatin capsule, background.\\[3pt]
\textbf{capsule2\_capsule3.} F: black gelatin capsule, orange gelatin capsule. B: white printed number 500, background.
\end{promptbox}

\begin{promptbox}
\textbf{Hazelnut.}\\[2pt]
\textbf{hazelnut1.} F: hazelnut shell surface. B: hilum scar on hazelnut, background.\\[3pt]
\textbf{hazelnut2.} F: hilum scar on hazelnut. B: hazelnut shell surface, background.\\[3pt]
\textbf{hazelnut1\_hazelnut2.} F: hazelnut shell surface, hilum scar on hazelnut. B: background.
\end{promptbox}

\begin{promptbox}
\textbf{Metal nut.}\\[2pt]
\textbf{metal\_nut1.} F: outer hexagonal metal body. B: central threaded hole, background.\\[3pt]
\textbf{metal\_nut2.} F: central threaded hole. B: outer hexagonal metal body, background.\\[3pt]
\textbf{metal\_nut1\_metal\_nut2.} F: outer hexagonal metal body, central threaded hole. B: background.
\end{promptbox}

\begin{promptbox}
\textbf{Pill.}\\[2pt]
\textbf{pill1.} F: plain pill surface. B: embossed letters FF on pill, background.\\[3pt]
\textbf{pill2.} F: embossed letters FF on pill. B: plain pill surface, background.\\[3pt]
\textbf{pill1\_pill2.} F: plain pill surface, embossed letters FF on pill. B: background.
\end{promptbox}

\begin{promptbox}
\textbf{Screw.}\\[2pt]
\textbf{screw1.} F: threaded screw shaft. B: screw head top, screw neck, background.\\[3pt]
\textbf{screw2.} F: screw head top. B: threaded screw shaft, screw neck, background.\\[3pt]
\textbf{screw3.} F: screw neck. B: threaded screw shaft, screw head top, background.
\end{promptbox}

\begin{promptbox}
\textbf{Toothbrush.}\\[2pt]
\textbf{toothbrush1.} F: plastic toothbrush handle. B: white toothbrush bristles, colored toothbrush bristles, background.\\[3pt]
\textbf{toothbrush2.} F: white toothbrush bristles. B: plastic toothbrush handle, colored toothbrush bristles, background.\\[3pt]
\textbf{toothbrush3.} F: colored toothbrush bristles. B: plastic toothbrush handle, white toothbrush bristles, background.\\[3pt]
\textbf{toothbrush1\_toothbrush2.} F: plastic toothbrush handle, white toothbrush bristles. B: colored toothbrush bristles, background.\\[3pt]
\textbf{toothbrush2\_toothbrush3.} F: white toothbrush bristles, colored toothbrush bristles. B: plastic toothbrush handle, background.\\[3pt]
\textbf{toothbrush1\_toothbrush2\_toothbrush3.} F: plastic toothbrush handle, white toothbrush bristles, colored toothbrush bristles. B: background.
\end{promptbox}

\begin{promptbox}
\textbf{Transistor.}\\[2pt]
\textbf{transistor1.} F: black plastic transistor body. B: metal transistor legs, background.\\[3pt]
\textbf{transistor2.} F: metal transistor legs. B: black plastic transistor body, background.\\[3pt]
\textbf{transistor1\_transistor2.} F: black plastic transistor body, metal transistor legs. B: background.
\end{promptbox}

\begin{promptbox}
\textbf{Zipper.}\\[2pt]
\textbf{zipper1.} F: interior zipper fabric. B: metal zipper teeth, fabric at zipper edge, background.\\[3pt]
\textbf{zipper2.} F: metal zipper teeth. B: interior zipper fabric, fabric at zipper edge, background.\\[3pt]
\textbf{zipper3.} F: fabric at zipper edge. B: interior zipper fabric, metal zipper teeth, background.\\[3pt]
\textbf{zipper1\_zipper2.} F: interior zipper fabric, metal zipper teeth. B: fabric at zipper edge, background.\\[3pt]
\textbf{zipper1\_zipper3.} F: interior zipper fabric, fabric at zipper edge. B: metal zipper teeth, background.\\[3pt]
\textbf{zipper2\_zipper3.} F: metal zipper teeth, fabric at zipper edge. B: interior zipper fabric, background.
\end{promptbox}

\subsection{AA-CLIP}
\label{app:prompts_scen2:aaclip}
AA-CLIP assigns a dedicated keyword to each subclass, naming the instructed component(s); multi-part subclasses combine the components with the conjunction ``AND''. The keyword is inserted into the same normal and abnormal state templates used in Scenario~1. The per-subclass keywords are listed below, one box per class.

\begin{promptbox}
\textbf{Cable.}\\[2pt]
\textbf{cable1:} light-blue insulation layer.\\[3pt]
\textbf{cable2:} green cable.\\[3pt]
\textbf{cable3:} blue cable.\\[3pt]
\textbf{cable4:} gray cable.\\[3pt]
\textbf{cable1\_cable2:} light-blue insulation layer AND green cable.\\[3pt]
\textbf{cable1\_cable3:} light-blue insulation layer AND blue cable.\\[3pt]
\textbf{cable2\_cable3:} green cable AND blue cable.\\[3pt]
\textbf{cable2\_cable4:} green cable AND gray cable.\\[3pt]
\textbf{cable3\_cable4:} blue cable AND gray cable.\\[3pt]
\textbf{cable2\_cable3\_cable4:} green cable AND blue cable AND gray cable.
\end{promptbox}

\begin{promptbox}
\textbf{Capsule.}\\[2pt]
\textbf{capsule1:} 500 white writing.\\[3pt]
\textbf{capsule2:} left black part.\\[3pt]
\textbf{capsule3:} right reddish-orange part.\\[3pt]
\textbf{capsule1\_capsule3:} 500 white writing AND right reddish-orange part.\\[3pt]
\textbf{capsule2\_capsule3:} left black part AND right reddish-orange part.
\end{promptbox}

\begin{promptbox}
\textbf{Hazelnut.}\\[2pt]
\textbf{hazelnut1:} shell.\\[3pt]
\textbf{hazelnut2:} hilum.\\[3pt]
\textbf{hazelnut1\_hazelnut2:} shell AND hilum.
\end{promptbox}

\begin{promptbox}
\textbf{Metal nut.}\\[2pt]
\textbf{metal\_nut1:} outer hexagonal body.\\[3pt]
\textbf{metal\_nut2:} central threaded hole.\\[3pt]
\textbf{metal\_nut1\_metal\_nut2:} outer hexagonal body AND central threaded hole.
\end{promptbox}

\begin{promptbox}
\textbf{Pill.}\\[2pt]
\textbf{pill1:} plain surface.\\[3pt]
\textbf{pill2:} embossed letters ``FF''.\\[3pt]
\textbf{pill1\_pill2:} plain surface AND embossed letters ``FF''.
\end{promptbox}

\begin{promptbox}
\textbf{Screw.}\\[2pt]
\textbf{screw1:} threaded shaft.\\[3pt]
\textbf{screw2:} screw head.\\[3pt]
\textbf{screw3:} neck.
\end{promptbox}

\begin{promptbox}
\textbf{Toothbrush.}\\[2pt]
\textbf{toothbrush1:} plastic base.\\[3pt]
\textbf{toothbrush2:} white bristles.\\[3pt]
\textbf{toothbrush3:} colored bristles.\\[3pt]
\textbf{toothbrush1\_toothbrush2:} plastic base AND white bristles.\\[3pt]
\textbf{toothbrush2\_toothbrush3:} white bristles AND colored bristles.\\[3pt]
\textbf{toothbrush1\_toothbrush2\_toothbrush3:} plastic base AND white bristles AND colored bristles.
\end{promptbox}

\begin{promptbox}
\textbf{Transistor.}\\[2pt]
\textbf{transistor1:} plastic body.\\[3pt]
\textbf{transistor2:} metalic legs.\\[3pt]
\textbf{transistor1\_transistor2:} plastic body AND metalic legs.
\end{promptbox}

\begin{promptbox}
\textbf{Zipper.}\\[2pt]
\textbf{zipper1:} interior fabric.\\[3pt]
\textbf{zipper2:} zipper teeth.\\[3pt]
\textbf{zipper3:} fabric at the edge.\\[3pt]
\textbf{zipper1\_zipper2:} interior fabric AND zipper teeth.\\[3pt]
\textbf{zipper1\_zipper3:} interior fabric AND fabric at the edge.\\[3pt]
\textbf{zipper2\_zipper3:} zipper teeth AND fabric at the edge.
\end{promptbox}

\section{Scenario 3 Prompts}
\label{app:prompts_scen3}
This appendix reports the textual inputs used in the Scenario~3 (Assembled Panel) evaluation (Section~\ref{sec:res_scen3}), organized by model and split into the two evaluation modes: Standard~AD (one prompt for all categories) and Per-Error~AD (one prompt per anomaly type, naming the relevant component). Anomaly types follow the taxonomy of Table~\ref{tab:ap_taxonomy} (identifiers A1--A14).

\subsection{AnomalyGPT}
\label{app:prompts_scen3:anomalygpt}

\paragraph{Standard AD.}
\label{app:prompts:scenario3_std}
Four prompts of increasing descriptive richness are evaluated, the same prompt for all images.

\begin{promptbox}
\textbf{Minimal.}~``Is there an error in the image?'' \\[3pt]
\textbf{Simple.}~``This is a photo of an electrical panel for anomaly detection, which should be complete and intact, without any damage, flaw, defect, scratch, hole or broken part. Is there any anomaly in the image?'' \\[3pt]
\textbf{Detailed.}~``This is a photo of an electrical panel for anomaly detection. It should contain two symmetrical circuit boards on either end, with three relays, properly terminated wiring, and no disconnected or floating wires. The wiring should be clean, secured with clips or zip ties, and each cable should be fully connected. Is there any anomaly in the image?'' \\[3pt]
\textbf{Very Detailed.}~``This is a photo of an electrical control panel designed for industrial applications. The panel should contain two identical green printed circuit boards mounted symmetrically at both ends. Each board must include three white relays aligned properly, labeled with part numbers, and connected to orange terminal blocks. The wiring should be cleanly routed using multicolored cables (e.g., yellow, brown, blue, purple, orange, and white), all securely connected to their respective terminals. In the central section, there should be two identical electromechanical actuators (such as contactors or switches), each with a red safety latch. The wiring should be neatly bundled and held down with zip ties and cable clips. All cables must be terminated properly; no wires should be loose, floating, or disconnected. Additionally, a flat ribbon cable may be present for communication between components, and all wiring should be consistent with standard color codes and labeling (e.g., L1, L2, L3). The panel should not have any burn marks, corrosion, broken solder joints, missing components, or any visual signs of damage. Is there any anomaly in the image?''
\end{promptbox}

\paragraph{Per-Error AD.} One natural-language prompt per anomaly type, naming the relevant component and the expected failure mode.

\begin{promptbox}
\textbf{A1 (swapped power-cable colors).}\\[2pt]
``This is a photo of an electrical panel for anomaly detection, which should be complete and intact, without swapped colors in power cables. Cables of different colors (e.g., blue and brown) should not be swapped at the left and right power ports. The correct ordering of cables viewed from top to bottom is that the left port should have brown, blue, green while the right port should have green, blue, brown. Is there any anomaly in the image?''
\end{promptbox}

\begin{promptbox}
\textbf{A2 (missing label).}\\[2pt]
``This is a photo of an electrical panel for anomaly detection, which should be complete and intact, without a missing label. All required number labels must be present on top of the black box in the left and right PCB. Is there any anomaly in the image?''
\end{promptbox}

\begin{promptbox}
\textbf{A3 (missing rubber ring).}\\[2pt]
``This is a photo of an electrical panel for anomaly detection, which should be complete and intact, without a missing rubber ring. Rubber rings should be present on all screws that fasten the metal panel. Only the panel-mounting screws need to be inspected; PCB screws are not part of this check. Is there any anomaly in the image?''
\end{promptbox}

\begin{promptbox}
\textbf{A4 (missing power-connection wires).}\\[2pt]
``This is a photo of an electrical panel for anomaly detection, which should be complete and intact, without missing power connection wires. All expected power wires must be connected to the top and bottom ports. Is there any anomaly in the image?''
\end{promptbox}

\begin{promptbox}
\textbf{A5 (missing screws).}\\[2pt]
``This is a photo of an electrical panel for anomaly detection, which should be complete and intact, without missing screws. All screws must be in place on the panel and PCB. Is there any anomaly in the image?''
\end{promptbox}

\begin{promptbox}
\textbf{A6 (incorrect wire routing through current sensor).}\\[2pt]
``This is a photo of an electrical panel for anomaly detection, which should be complete and intact, without incorrect wire routing through the current sensor on the right. In the correct configuration, exactly two wires, the brown and the blue power cables, must pass through the right current sensor. These two cables should be routed together through the sensor opening, and no additional or missing wires should occur. Is there any anomaly in the image?''
\end{promptbox}

\begin{promptbox}
\textbf{A7 (scratched or broken PCB).}\\[2pt]
``This is a photo of an electrical panel for anomaly detection, which should be complete and intact, without a scratched or broken PCB. The board should not have coating damage, scratches, broken components, or missing components. Is there any anomaly in the image?''
\end{promptbox}

\begin{promptbox}
\textbf{A8 (swapped large EV-port cables).}\\[2pt]
``This is a photo of an electrical panel for anomaly detection, which should be complete and intact, without swapped large cables in the EV port. Green and blue cables must not be switched between ports. The correct ordering viewed from top to bottom is: brown, green, blue on both ports. Is there any anomaly in the image?''
\end{promptbox}

\begin{promptbox}
\textbf{A9 (swapped small EV-port cables).}\\[2pt]
``This is a photo of an electrical panel for anomaly detection, which should be complete and intact, without swapped small cables in the EV port. Orange and white cables must not be switched between ports. The correct ordering viewed from top to bottom is: white, orange. Is there any anomaly in the image?''
\end{promptbox}

\begin{promptbox}
\textbf{A10 (swapped same-color IO power cables).}\\[2pt]
``This is a photo of an electrical panel for anomaly detection, which should be complete and intact, without swapped same-color upper and lower IO power cables. Both cables, oriented toward the left side of the image for the left PCB, must be correctly connected to their corresponding PCB terminal blocks (orange power terminal blocks). The cables must remain in their designated positions and must not be swapped between the upper and lower power ports. Is there any anomaly in the image?''
\end{promptbox}

\begin{promptbox}
\textbf{A11 (misplaced label).}\\[2pt]
``This is a photo of an electrical panel for anomaly detection, which should be complete and intact, without a misplaced label. Number labels should be correctly placed on top of the black box in the left and right PCB. Is there any anomaly in the image?''
\end{promptbox}

\begin{promptbox}
\textbf{A12 (missing red cap).}\\[2pt]
``This is a photo of an electrical panel for anomaly detection, which should be complete and intact, without a missing red cap. The red protective cap near the EV port should always be attached. Is there any anomaly in the image?''
\end{promptbox}

\begin{promptbox}
\textbf{A13 (missing EV-port wires).}\\[2pt]
``This is a photo of an electrical panel for anomaly detection, which should be complete and intact, without missing wires on EV port connections. All EV port wires should be fully connected. Is there any anomaly in the image?''
\end{promptbox}

\begin{promptbox}
\textbf{A14 (unplugged wires on current sensor).}\\[2pt]
``This is a photo of an electrical panel for anomaly detection, which should be complete and intact, without unplugged wires on the current sensor. All wires passing through the right current sensor must be fully connected. Is there any anomaly in the image?''
\end{promptbox}

\subsection{LogSAD}
\label{app:prompts_scen3:logsad}

\paragraph{Standard AD.} A foreground keyword and a background keyword are inserted into the template ensemble. Three foreground keywords are evaluated against two background keywords, the same configuration for all images. The following is the list, ordered from Minimal to Very Detailed.

\begin{promptbox}
\textbf{Foreground keywords.}~electrical control panel; industrial PCB assembly; industrial electrical control panel with green PCBs, white relays, orange terminals, and multicolored cables. \\[3pt]
\textbf{Background keywords.}~background; industrial environment.
\end{promptbox}

\paragraph{Per-Error AD.} Per-anomaly-type foreground (F) and background (B) keyword sets.

\begin{promptbox}
\textbf{Per-Error keyword sets.}\\[2pt]
\textbf{A1} (swapped power-cable colors). F: left and right power ports; focus on cable colors and positions. B: general PCB surface and other components.\\[3pt]
\textbf{A2} (missing label). F: top of left and right PCB; focus on missing number labels. B: general PCB surface.\\[3pt]
\textbf{A3} (missing rubber ring). F: panel-mounting screws on the metal panel; focus on rubber rings. B: other screws and unrelated PCB areas.\\[3pt]
\textbf{A4} (missing power-connection wires). F: top and bottom power ports; focus on missing power wires. B: general PCB area outside these ports.\\[3pt]
\textbf{A5} (missing screws). F: panel and PCB screws; focus on missing fasteners. B: panel borders and non-screw regions.\\[3pt]
\textbf{A6} (incorrect wire routing through current sensor). F: right current sensor with the brown and blue power cables; focus on routing through the sensor. B: PCB areas unrelated to the sensor.\\[3pt]
\textbf{A7} (scratched or broken PCB). F: damaged PCB regions; focus on scratches, cracks, or missing components. B: undamaged PCB and panel areas.\\[3pt]
\textbf{A8} (swapped large EV-port cables). F: left and right EV ports; focus on green, blue, and brown cables. B: general PCB surface and non-EV-port areas.\\[3pt]
\textbf{A9} (swapped small EV-port cables). F: left and right EV ports; focus on orange and white cables. B: general PCB surface and surrounding areas.\\[3pt]
\textbf{A10} (swapped same-color IO power cables). F: left and right IO power cables; focus on upper and lower positions on the terminal blocks. B: surrounding PCB areas.\\[3pt]
\textbf{A11} (misplaced label). F: black-box areas on left and right PCB; focus on correct label placement on top. B: general PCB surface and panel.\\[3pt]
\textbf{A12} (missing red cap). F: area near the EV port; focus on the red protective cap. B: general PCB and non-EV-port areas.\\[3pt]
\textbf{A13} (missing EV-port wires). F: left and right EV ports; focus on missing EV-port wires. B: other PCB areas and ports.\\[3pt]
\textbf{A14} (unplugged wires on current sensor). F: right current sensor; focus on disconnected or unplugged wires. B: PCB areas unrelated to the sensor.
\end{promptbox}

\subsection{AA-CLIP}
\label{app:prompts_scen3:aaclip}

\paragraph{Standard AD.} Three foreground keywords are evaluated, the same keyword for all images. The following is the list, ordered from Minimal to Very Detailed.

\begin{promptbox}
\textbf{Foreground keywords.}~electrical control panel; industrial PCB assembly; industrial electrical control panel with green PCBs, white relays, orange terminals, and multicolored cables.
\end{promptbox}

\paragraph{Per-Error AD.} One component-focused keyword per anomaly type, inserted into the state-conditioned templates.

\begin{promptbox}
\textbf{Per-Error keywords.}\\[2pt]
\textbf{A1} (swapped power-cable colors): left and right power ports; cable colors and positions.\\[3pt]
\textbf{A2} (missing label): top of left and right PCB; missing number labels.\\[3pt]
\textbf{A3} (missing rubber ring): panel-mounting screws; rubber rings.\\[3pt]
\textbf{A4} (missing power-connection wires): top and bottom power ports; missing power wires.\\[3pt]
\textbf{A5} (missing screws): panel and PCB screws; missing fasteners.\\[3pt]
\textbf{A6} (incorrect wire routing through current sensor): right current sensor; routing of the brown and blue cables.\\[3pt]
\textbf{A7} (scratched or broken PCB): damaged PCB regions; scratches, cracks, or missing components.\\[3pt]
\textbf{A8} (swapped large EV-port cables): left and right EV ports; green, blue, and brown cables.\\[3pt]
\textbf{A9} (swapped small EV-port cables): left and right EV ports; orange and white cables.\\[3pt]
\textbf{A10} (swapped same-color IO power cables): left and right IO power cables; upper and lower terminal-block positions.\\[3pt]
\textbf{A11} (misplaced label): black-box areas on left and right PCB; label placement.\\[3pt]
\textbf{A12} (missing red cap): area near the EV port; red protective cap.\\[3pt]
\textbf{A13} (missing EV-port wires): left and right EV ports; missing EV-port wires.\\[3pt]
\textbf{A14} (unplugged wires on current sensor): right current sensor; unplugged wires.
\end{promptbox}

\end{document}